\documentclass[10pt,twocolumn,letterpaper]{article}

\usepackage[pagenumbers]{cvpr} 

\usepackage{graphicx}
\usepackage{amsmath}
\usepackage{amssymb}
\usepackage{booktabs}
\usepackage{multirow}
\usepackage{cases}
\usepackage{mathtools}
\usepackage{algorithm}
\usepackage{algpseudocode}

\definecolor{cvprblue}{rgb}{0.21,0.49,0.74}
\usepackage[pagebackref,breaklinks,colorlinks,allcolors=cvprblue]{hyperref}

\def\confName{CVPR}
\def\confYear{2027}

\title{SEEL: Towards End-to-End 3D Sign Generation with Language Models}

\author{Ronglai Zuo \qquad Fei Ni \qquad Junpeng Jing \qquad Rolandos Alexandros Potamias \\ Jiankang Deng \qquad Stefanos Zafeiriou\\
Imperial College London \\
{\tt\small \href{https://2000zrl.github.io/seel/}{https://2000zrl.github.io/seel/}}
}

\begin{document}
\twocolumn[{
\maketitle
\begin{center}
\centering
\includegraphics[width=1.0\textwidth]{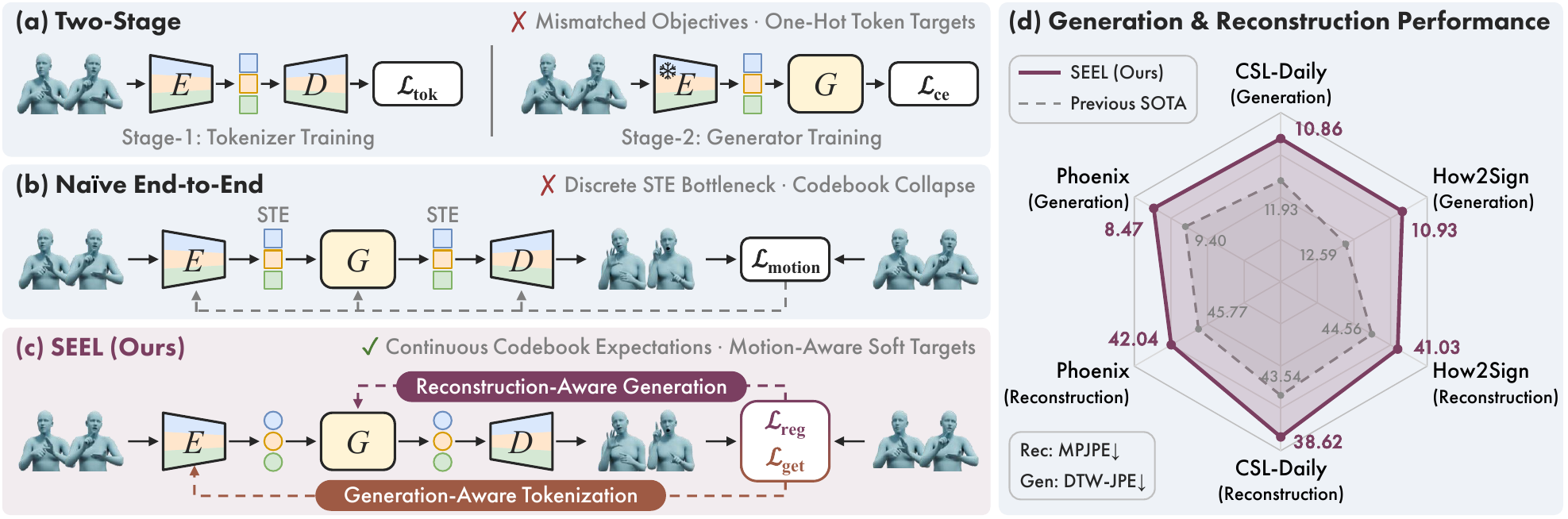}
\vspace{-6mm}
\captionof{figure}{\textbf{Paradigms of 3D Sign Language Generation.}
(a) Two-stage training isolates the frozen tokenizer (encoder $E$ and decoder $D$) from generation feedback and supervises the generator $G$ with one-hot token targets.
(b) Na\"ive end-to-end training relies on straight-through estimators (STE)~\cite{bengio2013ste,vqvae}, suffering from biased gradients and codebook collapse.
(c) SEEL decomposes generated-motion supervision ($\mathcal{L}_{\mathrm{motion}}$) via two complementary mechanisms, bypassing STE across both $E\rightarrow G$ and $G\rightarrow D$ interfaces.
(d) Both the generation and reconstruction performance are improved over the previous SOTA LM-based method~\cite{zuo2026madis} on three benchmarks~\cite{csl-daily,duarte2021how2sign,2014T}.}
\label{fig:teaser}
\end{center}
}]

\begin{abstract}
3D sign language generation aims to produce signing motion from written text. Motivated by the linguistic nature of sign language, recent methods combine a motion tokenizer with a language model to predict motion tokens from text. However, both components are optimized through a two-stage pipeline with mismatched objectives: the tokenizer minimizes reconstruction error in isolation, while the generator learns from rigid one-hot reference tokens. A straightforward end-to-end extension bridges discrete token selection with straight-through estimators (STE), suffering from biased gradients and codebook collapse. To address these limitations, we propose SEEL, an end-to-end framework that aligns tokenization and generation with motion quality through two complementary mechanisms. First, to overcome rigid one-hot supervision, reconstruction-aware generation decodes candidate tokens in context and converts their 3D motion errors into soft targets, guiding the generator with physical motion quality. Second, to propagate generation feedback without STE, generation-aware tokenization passes continuous codebook expectations through the frozen generator, routing generated-motion gradients to the tokenizer to prevent codebook collapse. Combining both mechanisms enables the tokenizer and generator to co-evolve, achieving state-of-the-art performance on CSL-Daily, How2Sign, and Phoenix-2014T. Code and models will be released to facilitate future research.
\end{abstract}

\vspace{-5mm}
\section{Introduction}
\label{sec:intro}

3D sign language generation (SLG) aims to translate written text into expressive signing motion, serving as a vital bridge between deaf and hearing communities~\cite{baltatzis2024nsa,lee2026signer,bensabath2025text,low2026signspark,zuo2025soke,yu2025signavatars,liu2026bridging,forte2023reconstructing}.
Inspired by the linguistic nature of sign language~\cite{zuo2025soke,yin2021including} and the potential of unified modeling of signs and texts~\cite{liu2026bridging}, language-model-based (LM-based) paradigms have emerged as a promising approach for SLG~\cite{zuo2025soke,yin-etal-2024-t2s,zuo2026madis,yu2025signavatars,fang2025signllm,fg2024signavatar}.
These methods typically follow a two-stage pipeline: a motion tokenizer~\cite{zuo2025soke,vqvae} first quantizes continuous 3D motion into discrete tokens, and a language model~\cite{yang2025qwen3,mbart} subsequently learns to predict token sequences conditioned on text.

Despite its strong modeling capacity, decoupling tokenization from generation introduces a fundamental objective mismatch across the pipeline (\cref{fig:teaser}a). 
Because the tokenizer is optimized solely for motion reconstruction while the language model is trained to match one-hot reference tokens, high reconstruction fidelity does not necessarily translate into high generation quality~\cite{vqgan,yu2024magvit,ramanujan2026worse}. Specifically, isolated tokenizer training yields latent spaces that remain unadapted to text-conditioned generation, while one-hot token supervision penalizes all non-reference tokens equally regardless of their decoded motion quality.
Although connecting both stages end-to-end via straight-through estimators (STE)~\cite{vqvae,bengio2013ste} is a natural remedy (\cref{fig:teaser}b), discrete selection in LM-based generation occurs across two interfaces, \ie, quantizing encoder features into tokens for the generator and selecting predicted tokens for the decoder. 
Applying STE across both interfaces introduces gradient bias and causes codebook collapse~\cite{huh2023straightening, yin2019understanding}, which is particularly detrimental to SLG given its high demand for fine-grained hand gestures and finger articulations.

To overcome these challenges, we propose SEEL, an end-to-end training framework that aligns both tokenization and generation with downstream motion quality without relying on STE (\cref{fig:teaser}c). Built upon a vector-quantized motion tokenizer~\cite{zuo2025soke} and a masked diffusion language model~\cite{zuo2026madis,llada} that share a motion decoder, SEEL establishes STE-free, motion-aware supervision across the tokenization-generation interface through two complementary mechanisms: reconstruction-aware generation (REG) and generation-aware tokenization (GET).

Specifically, REG complements rigid one-hot labels with a soft target distribution that directly reflects 3D motion quality. While recent works incorporate 3D supervision via physical-space pretraining~\cite{zuo2026madis} or reconstruction regularization~\cite{low2026signspark}, they formulate motion-aware supervision by regressing a single decoded motion sequence. For supervising an LM-based generator, however, regressing a single motion sequence fails to inform the model how alternative tokens in the codebook compare in motion space. Ideally, capturing these comparisons would require evaluating every token in the codebook at each masked position, but exhaustive decoding is computationally infeasible. To make this tractable, REG selects a compact, informative set of candidate tokens and decodes them in context through the motion decoder to measure their reconstruction errors. Converting these errors into a soft target distribution explicitly guides the generator towards tokens with high 3D motion fidelity rather than enforcing a strict index match.

Complementarily, GET extends motion-aware supervision to the tokenizer, directly optimizing the motion encoder with downstream generation feedback. To bypass the non-differentiable hard token selection of STE, GET replaces discrete indexing with continuous expectations over the codebook. At unmasked context positions, soft assignments computed from the distances between encoded motion features and codebook entries produce continuous context embeddings for the frozen generator. At masked prediction positions, the generator's output distributions compute expected code embeddings, which are combined with unmasked reference tokens and decoded through the frozen motion decoder into 3D motion. This formulation creates a fully differentiable pathway that routes gradients from generated-motion errors back to the encoder, preventing codebook collapse and regularizing the latent space towards token representations predictable by the generator.

By jointly aligning the generator with 3D motion fidelity and optimizing the tokenizer with downstream generation feedback, SEEL achieves a mutually reinforcing synergy that simultaneously improves both generation and reconstruction performance (\cref{fig:teaser}d). Crucially, both REG and GET operate exclusively as training objectives, introducing zero architectural overhead or latency at inference. 
Our main contributions are summarized as follows:
\begin{itemize}
    \item We propose SEEL, the first end-to-end training framework for LM-based 3D SLG, which bridges the objective mismatch of two-stage pipelines while circumventing the gradient bias and codebook collapse of STE.
    \item We introduce reconstruction-aware generation, which decodes a compact set of candidate tokens in context and converts their 3D motion errors into soft target distributions, guiding the generator with 3D motion fidelity beyond rigid one-hot supervision.
    \item We design generation-aware tokenization, which replaces hard token selection at both quantization interfaces with continuous codebook expectations, creating a differentiable pathway that optimizes the motion encoder directly with generated-motion errors.
    \item Extensive experiments on three widely-adopted benchmarks~\cite{csl-daily,duarte2021how2sign,2014T} demonstrate that SEEL achieves state-of-the-art performance in 3D sign language generation.
\end{itemize}

\begin{figure*}[t]
\centering
\includegraphics[width=1.0\linewidth]{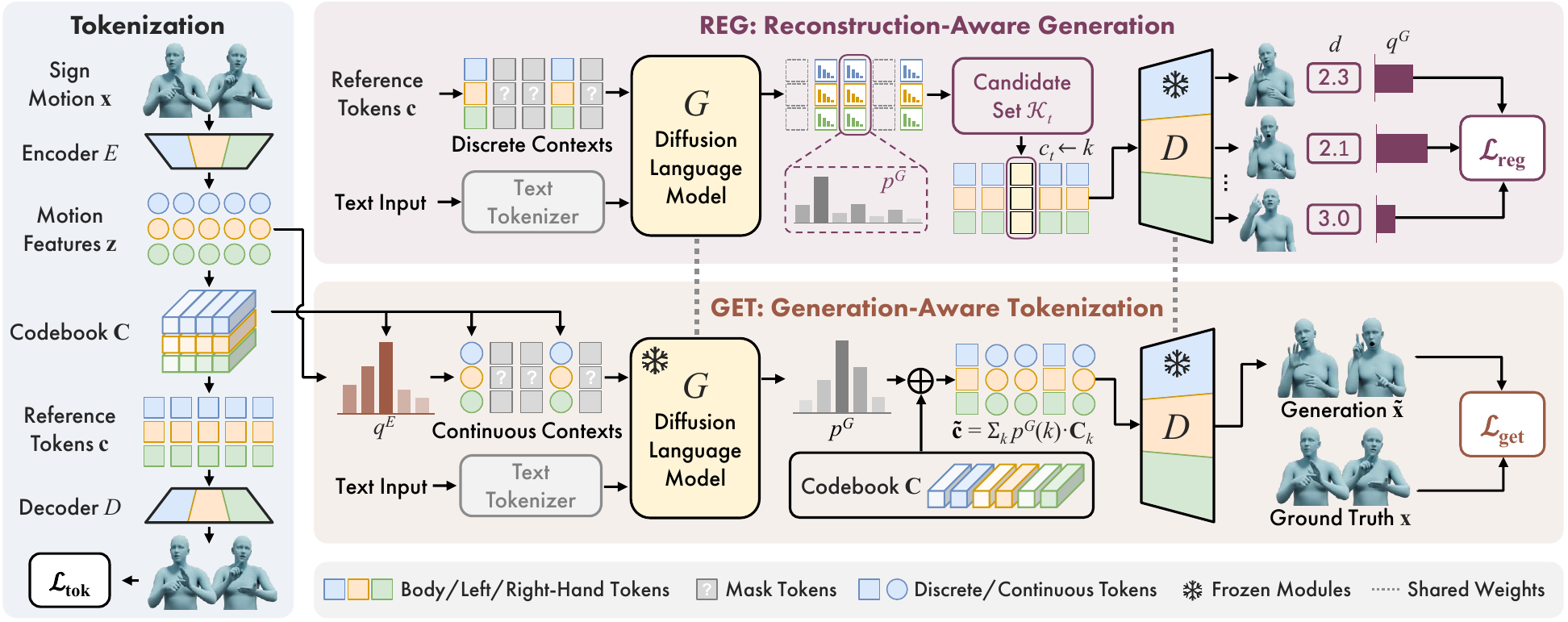}
\vspace{-7mm}
\caption{\textbf{Overview of SEEL.}
\textbf{Tokenization (Left):} The motion encoder maps sign motion $\mathbf{x}$ to latent features $\mathbf{z}$, which are quantized into discrete reference tokens $\mathbf{c}$ and decoded to reconstruct the motion.
\textbf{REG (Top):} Given text and discrete contexts, the generator predicts token distributions $p^G$ at masked positions. Substituting candidate tokens ($c_t\leftarrow k \in \mathcal{K}_t$) and decoding them in context yields motion errors $d$, which are converted into soft targets $q^G$ to guide generator training.
\textbf{GET (Bottom):} Soft codebook assignments $q^E$ provide continuous context embeddings to the frozen generator. Its predictions form expected code embeddings $\tilde{\mathbf{c}}$, which are decoded into generated motion $\tilde{\mathbf{x}}$ to train the encoder without STE.}
\vspace{-3mm}
\label{fig:framework}
\end{figure*}
\section{Related Work}
\noindent\textbf{Sign Language Generation.}
Sign language generation (SLG), or sign language production, translates text into sign language representations, including skeletons~\cite{tang2025discrete,tang2025sign,liu2026signpr,yu2026focal,saunders2020progressive,arkushin2023ham2pose,saunders2021mixed}, RGB videos~\cite{qi2026signmod,wang2026advanced,fang2026stable,saunders2022signing,pelykh2026signrefine,signgen}, and 3D motions~\cite{zuo2025soke,low2026signspark,yu2025signavatars,yin-etal-2024-t2s,baltatzis2024nsa,lee2026signer,xia2026signmask}. Early methods rely on glosses as intermediate queries to retrieve and stitch individual signs~\cite{zuo2024simple,saunders2022signing,saunders2020progressive,stoll2020text2sign,huang2022dualsign,tang2022gloss,fang2026stable,lee2026signer} or as auxiliary supervision~\cite{saunders2021mixed}, but gloss annotations are costly and suffer from information loss~\cite{muller2022considerations}. Consequently, recent works have shifted towards gloss-free SLG. While some approaches rely on continuous visual generative models~\cite{qi2026signmod,baltatzis2024nsa,ye2026hybrid,wang2026advanced,low2026signspark}, others draw on the linguistic nature of sign language by combining a discrete motion tokenizer with a language model~\cite{yu2025signavatars,fg2024signavatar,zuo2025soke,yin-etal-2024-t2s,zuo2026madis}, demonstrating strong potential for unifying sign language understanding and generation~\cite{liu2026bridging,an2026signflip,guo2024unsupervised}. 

In this work, we focus on gloss-free 3D sign generation within the language-model paradigm, producing motions that can drive downstream video synthesis~\cite{shi2024pose,Shi_2025_ICCV}, avatar renderers~\cite{zhang2025guava,wang2026phosa,ivashechkin2025signsplat,lakhal2025diverse}, and humanoid robot control~\cite{liu2026traceable,qiao2026signbot,khan2026sign}. Unlike prior two-stage pipelines with a frozen tokenizer, we present the first end-to-end framework that jointly optimizes the motion tokenizer and generator.

\noindent\textbf{End-to-End Visual Generation.}
Modern generative models across images, videos, and 3D human motions span latent diffusion over continuous variational autoencoders (VAEs)~\cite{latentdiff,peebles2023scalable,blattmann2023align,chen2023executing} and autoregressive or masked models over discrete tokenizers~\cite{vqvae,vqgan,chang2022maskgit,sun2024autoregressive,yu2024magvit,t2m-gpt,guo2024momask}. These approaches predominantly follow a two-stage paradigm, where the tokenizer is trained solely for reconstruction and frozen during generator training. Although recent works align latents with pretrained visual foundation models~\cite{yao2025reconstruction,yu2025repa,zhang2026videorepa}, co-train a video tokenizer with a lightweight autoregressive prior~\cite{wang2025larp}, or inject language guidance into motion tokenization~\cite{yan2026lgtok}, the full generator remains decoupled from tokenizer optimization. However, na\"ively unfreezing the tokenizer for joint training degrades performance, as directly backpropagating generative losses causes variance collapse in continuous VAEs~\cite{leng2025repae,zheng2026genfirst} and codebook collapse under straight-through estimation in discrete tokenizers~\cite{chu2026eostok}. To relieve these issues, recent works guide joint optimization through representation alignment with pretrained foundation models~\cite{leng2025repae,chu2026eostok}, explicit entropy regularization~\cite{zheng2026genfirst}, or staged training schedules~\cite{zheng2026genfirst}.

Unlike image and video synthesis, 3D sign language generation lacks off-the-shelf foundation models for external representation alignment. Instead, our proposed SEEL enables self-contained end-to-end training of the discrete tokenizer and LM-based generator by combining motion-aware soft targets with continuous codebook expectations.

\section{Methodology}
As shown in \cref{fig:framework}, SEEL jointly trains a motion tokenizer and a language-model generator. During training, REG decodes candidate tokens in context to guide the generator with motion-aware soft targets, while GET passes continuous codebook expectations through the frozen generator and decoder to update the encoder from generated-motion errors. At inference, the generator predicts discrete tokens from text, which the decoder converts into 3D sign motion.

\subsection{Preliminaries}
\noindent\textbf{LM-Based SLG} combines a vector-quantized VAE~\cite{vqvae} motion tokenizer with a language-model generator. In the tokenizer, a motion encoder $E$ maps a sign sequence $\mathbf{x}$ to $T$ latent features $\mathbf{z}=E(\mathbf{x})$. Given a codebook $\mathbf{C}=\{\mathbf{C}_k\}_{k=1}^{N}$ containing $N$ embeddings, each feature is assigned a token index $c_t = \arg\min_k \|\mathbf{z}_t - \mathbf{C}_k\|_2^2$. The resulting token sequence $\mathbf{c}$ is decoded into reconstructed motion $\hat{\mathbf{x}}=D(\mathbf{C}[\mathbf{c}])$, where $D$ is the motion decoder and $\mathbf{C}[\mathbf{c}]$ denotes the corresponding code embeddings. 
The tokenizer objective $\mathcal{L}_{\mathrm{tok}}$ combines a reconstruction loss $\|\hat{\mathbf{x}} - \mathbf{x}\|_1$ (where $\|\cdot\|_1$ denotes smooth $L_1$) with commitment regularization, while the codebook $\mathbf{C}$ is updated via exponential moving averages (EMA) of assigned encoder features~\cite{vqvae,jiang2023motiongpt,zuo2025soke}. Following~\cite{zuo2025soke}, the tokenizer models the body (including face), left hand, and right hand as three parallel streams. Since all tokenization and generation operations apply identically to each stream, we omit stream indices for simplicity. 

We instantiate the generator $G$ as a masked diffusion language model due to its superior performance and efficiency on SLG~\cite{zuo2026madis}. In conventional two-stage training, the tokenizer remains frozen while $G$ predicts masked motion tokens conditioned on written text $\mathbf{y}$ and unmasked token context $\mathbf{c}_{\mathrm{ctx}}$. Training minimizes $\mathcal{L}_{\mathrm{ce}}=-\frac{1}{|\mathcal{S}|} \sum_{t\in\mathcal{S}}\log p_t^G(c_t|\mathbf{y},\mathbf{c}_{\mathrm{ctx}})$, where $\mathcal{S}$ denotes the masked prediction positions and $p^G$ is the model's conditional token probability. More details are available in the appendix.

\begin{figure}[t]
\centering
\includegraphics[width=1.0\linewidth]{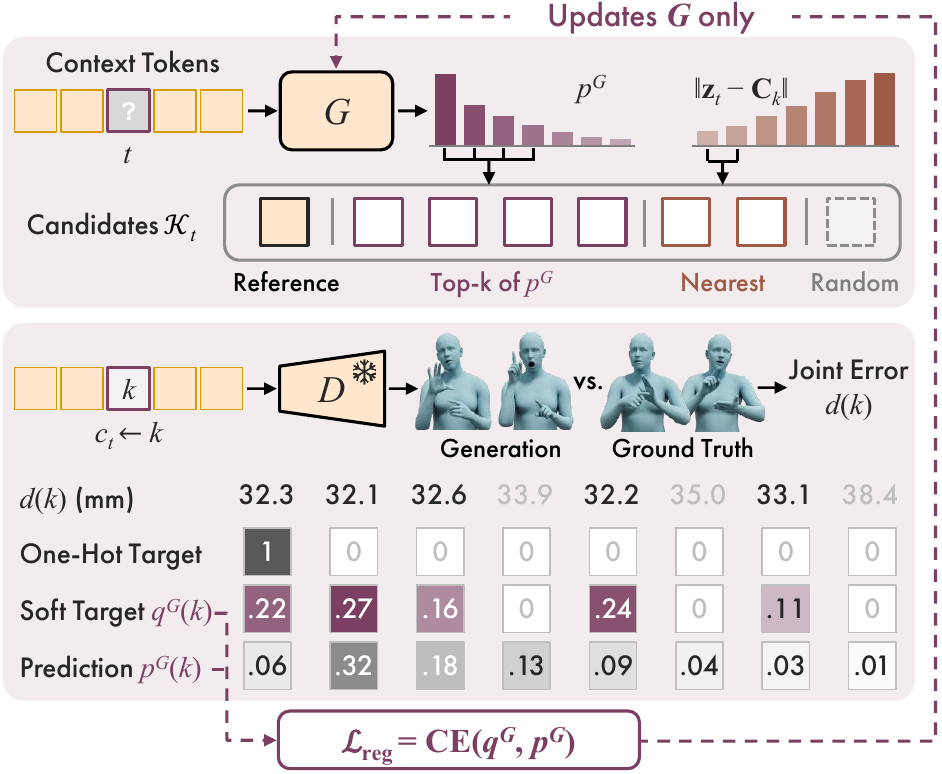}
\vspace{-6mm}
\caption{\textbf{Reconstruction-Aware Generation.}
At each masked position $t$, a candidate set $\mathcal{K}_t$ is constructed from four sources. Substituting each candidate in context and decoding the resulting sequence yields the motion error $d(k)$. After masking out poor candidates (grey entries), the remaining errors are converted into soft targets $q^G$ to supervise the generator.}
\vspace{-2mm}
\label{fig:reg}
\end{figure}

\noindent\textbf{A Na\"ive End-to-End Baseline.}
While $\mathcal{L}_{\mathrm{tok}}$ only supervises motion decoded from the encoder's tokens ($\hat{\mathbf{x}}$), optimizing $\mathcal{L}_{\mathrm{tok}} + \mathcal{L}_{\mathrm{ce}}$ concurrently leaves $E$ without generation feedback: $\mathcal{L}_{\mathrm{ce}}$ updates only $G$ against detached token indices ($\nabla_E \mathcal{L}_{\mathrm{ce}} = 0$) and never decodes $G$'s predictions into 3D motion. 
To introduce generation feedback, a straightforward end-to-end baseline decodes $G$'s predicted tokens into sign motion $\tilde{\mathbf{x}}$ and penalizes $\mathcal{L}_{\mathrm{motion}} = \|\tilde{\mathbf{x}} - \mathbf{x}\|_1$ (~\cref{fig:teaser}b).
Since hard token selection is non-differentiable, this baseline applies a straight-through estimator (STE)~\cite{vqvae,bengio2013ste} at both the generator-input and generator-output interfaces: $\mathbf{e}_{\mathrm{ste}} = \mathbf{e}_{\mathrm{soft}} + \mathrm{sg}(\mathbf{e}_{\mathrm{hard}} - \mathbf{e}_{\mathrm{soft}})$, where $\mathbf{e}_{\mathrm{hard}}$ denotes the discretely selected code embedding, $\mathbf{e}_{\mathrm{soft}}$ is its continuous differentiable approximation, and $\mathrm{sg}(\cdot)$ stops gradients. 
This preserves discrete tokens in the forward pass while routing surrogate gradients through $\mathbf{e}_{\mathrm{soft}}$ in the backward pass. The overall baseline objective is $\mathcal{L}_{\mathrm{base}} = \mathcal{L}_{\mathrm{tok}} + \mathcal{L}_{\mathrm{ce}} + \lambda_{\mathrm{motion}}\mathcal{L}_{\mathrm{motion}}$.

\subsection{Reconstruction-Aware Generation (REG)}
\label{sec:reg}
In the na\"ive baseline, supervising the generator with $\mathcal{L}_{\mathrm{ce}}$ and $\mathcal{L}_{\mathrm{motion}}$ presents two key limitations: $\mathcal{L}_{\mathrm{ce}}$ enforces a rigid one-hot match toward $c_t$ that penalizes all alternative tokens equally regardless of their decoded motion quality, while $\mathcal{L}_{\mathrm{motion}}$ regresses only a single motion sequence via biased STE gradients without comparing competing token choices. To directly supervise the generator's output distribution with 3D motion fidelity while circumventing STE, REG decodes candidate tokens in context and converts their reconstruction errors into a soft target distribution (\cref{fig:reg}).

\noindent\textbf{Candidate Proposal.}
Evaluating all $N$ codebook entries at every masked position is computationally expensive, as each candidate substitution requires decoding a full token sequence. To keep evaluation tractable, we construct a small candidate set $\mathcal{K}_t$ at each masked position $t\in\mathcal{S}$. As shown in \cref{fig:reg}, $\mathcal{K}_t$ always retains the reference token $c_t$ alongside three complementary groups: top-ranked tokens from $p^G_t$ to capture alternatives the generator is likely to produce, codebook entries closest to the encoder feature $\mathbf{z}_t$ to recover nearby codes the generator might overlook, and random codes to encourage exploration.

\noindent\textbf{Contextual Candidate Scoring.}
For each candidate $k\in\mathcal{K}_t$, we replace $c_t$ with $k$ to obtain a modified sequence $\mathbf{c}^{(t\leftarrow k)}$, keeping all other token positions fixed. The shared decoder $D$ evaluates the resulting motion:
\begin{equation}
    \tilde{\mathbf{x}}^{(t\leftarrow k)} = D(\mathbf{C}[\mathbf{c}^{(t\leftarrow k)}]), \,
    d_t(k) = \ell_{\mathrm{joint}}(\tilde{\mathbf{x}}^{(t\leftarrow k)}, \mathbf{x}),
\end{equation}
where $\ell_{\mathrm{joint}}$ measures pelvis-relative whole-body 3D joint error over a 16-frame local window. Fixing the surrounding context isolates the motion effect of substituting candidate $k$ at position $t$.

\noindent\textbf{Generator Supervision with Soft Targets.}
We convert the candidate errors across $\mathcal{K}_t$ into a soft target distribution:
\begin{equation}
    q^G_t(k) = \frac{\exp(-d_t(k)/\tau_G)}{\sum_{j\in\mathcal{K}_t}\exp(-d_t(j)/\tau_G)},
\end{equation}
where $\tau_G>0$ is a temperature parameter. To filter out poor candidates, any $k\in\mathcal{K}_t$ whose error exceeds the reference error $d_t(c_t)$ by more than a margin $\delta$ is masked out and assigned zero probability in $q^G_t(k)$ (grey entries in \cref{fig:reg}). Remaining candidates with lower motion error receive higher target probabilities. REG then minimizes the cross-entropy between $q^G_t$ and the generator's predicted distribution $p^G_t$ over all masked positions:
\begin{equation}
    \mathcal{L}_{\mathrm{reg}} = -\frac{1}{|\mathcal{S}|} \sum_{t\in\mathcal{S}} \sum_{k\in\mathcal{K}_t} \operatorname{sg}\!\left(q^G_t(k)\right) \log p^G_t(k).
\end{equation}
Since candidate scoring and target construction are performed without gradients, $\mathcal{L}_{\mathrm{reg}}$ only updates the generator.

\begin{figure}[t]
\centering
\includegraphics[width=1.0\linewidth]{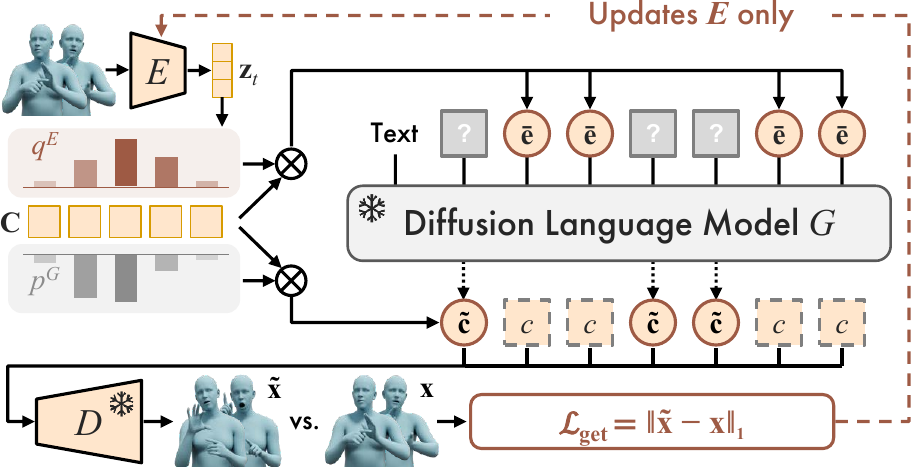}
\vspace{-6mm}
\caption{\textbf{Generation-Aware Tokenization.}
At unmasked positions, motion features $\mathbf{z}$ and codebook distances produce soft assignments $q^E$ to form continuous context embeddings $\bar{\mathbf{e}}$ for the frozen generator. At masked positions, predicted probabilities $p^G$ form expected code embeddings $\tilde{\mathbf{c}}$, which are combined with reference tokens $\mathbf{c}$ and decoded into motion $\tilde{\mathbf{x}}$ to train the encoder.}
\vspace{-2mm}
\label{fig:get}
\end{figure}

\subsection{Generation-Aware Tokenization (GET)}
\label{sec:get}
While REG aligns the generator with 3D motion quality, the motion encoder $E$ is still trained solely via self-reconstruction ($\mathcal{L}_{\mathrm{tok}}$) without feedback from downstream generation. In the na\"ive baseline, backpropagating generated-motion errors ($\mathcal{L}_{\mathrm{motion}}$) to $E$ must traverse hard token selections at both the generator output and input, compounding STE gradient bias across two quantization interfaces and inducing codebook collapse. To circumvent STE, GET replaces both discrete selections with continuous codebook expectations, enabling generated-motion gradients to directly optimize $E$ (\cref{fig:get}).

\noindent\textbf{Continuous Context Embeddings.}
To make the interface from $E$ to $G$ differentiable, we replace non-differentiable code lookup at unmasked context positions with a soft expectation over the codebook. Specifically, we convert the distances between the encoder feature $\mathbf{z}_t$ and codebook entries $\mathbf{C}_k$ into assignment probabilities $q^E_t(k)$ and compute the continuous context embedding $\bar{\mathbf{e}}_t$:
\begin{equation}
    \begin{gathered}
        q^E_t(k) = \frac{\exp(-\|\mathbf{z}_t-\operatorname{sg}(\mathbf{C}_k)\|_2^2 / \tau_E)}{\sum_{j=1}^{N}\exp(-\|\mathbf{z}_t-\operatorname{sg}(\mathbf{C}_j)\|_2^2/ \tau_E)}, \\
        \bar{\mathbf{e}}_t = \sum_{k=1}^{N} q^E_t(k)\operatorname{sg}(\mathbf{C}_k),
    \end{gathered}
\end{equation}
where $\tau_E>0$ is a temperature parameter. At unmasked positions, the generator receives $\bar{\mathbf{e}}_t$ in place of discrete context embeddings $\mathbf{C}[\mathbf{c}_{\mathrm{ctx}}]$, while masked positions retain their standard mask tokens.

\noindent\textbf{Continuous Predicted Embeddings.}
Conditioned on $\mathbf{y}$ and the continuous context embeddings, $G$ outputs the predicted token distribution $p^G_t(k)$ at each masked position $t\in\mathcal{S}$. Rather than selecting a single hard token, we compute the expected code embedding $\tilde{\mathbf{c}}_t$ over the predicted distribution and form the decoder input sequence $\mathbf{u}=(\mathbf{u}_1,\dots,\mathbf{u}_T)$:
\begin{equation}
    \tilde{\mathbf{c}}_t = \sum_{k=1}^{N} p^G_t(k)\operatorname{sg}(\mathbf{C}_k),
    \,
    \mathbf{u}_t =
    \begin{cases}
        \tilde{\mathbf{c}}_t, & t\in\mathcal{S},\\
        \operatorname{sg}(\mathbf{C}[c_t]), & t\notin\mathcal{S}.
    \end{cases}
\end{equation}
Here, masked prediction positions receive the continuous embeddings $\tilde{\mathbf{c}}_t$, whereas unmasked positions retain detached reference token embeddings $\operatorname{sg}(\mathbf{C}[c_t])$ to provide context for decoding.

\noindent\textbf{Encoder Supervision from Generation.}
The shared decoder $D$ maps $\mathbf{u}$ to a generated motion sequence $\tilde{\mathbf{x}}=D(\mathbf{u})$. GET then penalizes the motion error at the sequence level: $\mathcal{L}_{\mathrm{get}} = \|\tilde{\mathbf{x}}-\mathbf{x}\|_1.$
During this auxiliary computation, the parameters of $G$ and $D$ are frozen while their input gradients remain active, and therefore $\mathcal{L}_{\mathrm{get}}$ updates only the motion encoder $E$ via the unmasked context positions. 
Unlike STE, which routes biased identity gradients through a single selected code, $\mathcal{L}_{\mathrm{get}}$ propagates gradients across all codebook entries via this continuous expectation.

\subsection{Joint Optimization and Inference}
\label{sec:optim}
\noindent\textbf{Training Objective.}
Rather than backpropagating a single $\mathcal{L}_{\mathrm{motion}}$ term through STE as in $\mathcal{L}_{\mathrm{base}}$, SEEL decomposes generated-motion supervision into $\mathcal{L}_{\mathrm{reg}}$ and $\mathcal{L}_{\mathrm{get}}$, jointly optimizing the tokenizer and generator via:
\begin{equation}
    \mathcal{L}_{\mathrm{total}} = \mathcal{L}_{\mathrm{tok}} + \mathcal{L}_{\mathrm{ce}} + \lambda_{\mathrm{reg}}\mathcal{L}_{\mathrm{reg}} + \lambda_{\mathrm{get}}\mathcal{L}_{\mathrm{get}},
\end{equation}
where $\lambda_{\mathrm{reg}}$ and $\lambda_{\mathrm{get}}$ are loss weights. 
Within each training step, the four terms play complementary roles. 
$\mathcal{L}_{\mathrm{tok}}$ updates the encoder $E$ and shared decoder $D$ (with $\mathbf{C}$ updated via EMA) to maintain accurate motion reconstruction.
For the generator $G$, $\mathcal{L}_{\mathrm{ce}}$ learns the text-conditioned token distribution over the full codebook, while $\mathcal{L}_{\mathrm{reg}}$ locally refines this distribution within the candidate set $\mathcal{K}_t$ using decoded motion quality.
Finally, $\mathcal{L}_{\mathrm{get}}$ updates only the encoder $E$, encouraging it to produce token representations that $G$ can predict more accurately. 
Together, REG and GET align both tokenization and generation with downstream motion quality while eliminating biased STE gradients across both tokenization-generation interfaces ($E \rightarrow G$ and $G \rightarrow D$).

\noindent\textbf{Inference.}
REG and GET are training-only, introducing no test-time overhead. At inference, $E$ is discarded; given text $\mathbf{y}$, $G$ iteratively predicts discrete tokens $\hat{\mathbf{c}}$ from a fully masked sequence, and $D$ decodes $\mathbf{C}[\hat{\mathbf{c}}]$ into sign motion.

\begin{table*}[t]
\small
\setlength\tabcolsep{5pt}
    \centering
    \begin{tabular}{l|cccc|cccc|cccc}
    \toprule
    \multirow{3}{*}{Method} & \multicolumn{4}{c|}{CSL-Daily} & \multicolumn{4}{c|}{How2Sign} & \multicolumn{4}{c}{Phoenix-2014T} \\
    & \multicolumn{3}{c}{DTW-JPE$\downarrow$} & SiCLIP$\uparrow$ & \multicolumn{3}{c}{DTW-JPE$\downarrow$} & SiCLIP$\uparrow$ & \multicolumn{3}{c}{DTW-JPE$\downarrow$} & SiCLIP$\uparrow$ \\
    \cmidrule(r){2-4}\cmidrule(r){5-5}\cmidrule(r){6-8}\cmidrule(r){9-9}\cmidrule(r){10-12}\cmidrule(r){13-13}
    & All & Body & Hands & R@1 & All & Body & Hands & R@1 & All & Body & Hands & R@1 \\
    
    \midrule
    Prog. Trans.$^*$~\cite{saunders2020progressive} & 35.92 & 16.30 & 32.63 & 13.65 & 34.77 & 14.74 & 30.17 & 9.88 & 33.48 & 15.01 & 31.77 & 14.03 \\
    Text2Mesh$^*$~\cite{stoll2022there} & 33.63 & 13.76 & 30.37 & 15.84 & 35.79 & 15.50 & 32.97 & 9.30 & 33.05 & 14.04 & 31.64 & 12.08 \\
    T2S-GPT$^*$~\cite{yin-etal-2024-t2s} & 24.59 & 12.32 & 15.43 & 20.03 & 22.78 & 12.65 & 18.44 & 12.58 & 22.60 & 11.65 & 19.09 & 22.10 \\
    S-MotionGPT$^*$~\cite{jiang2023motiongpt} & 24.51 & 11.58 & 11.31 & 20.37 & 21.69 & 12.41 & 13.74 & 12.77 & 19.21 & 10.42 & 9.08 & 32.42 \\
    NSA \cite{baltatzis2024nsa} & -- & -- & -- & -- & 20.41 & 9.16 & 18.51 & 15.04 & -- & -- & -- & -- \\
    MoMask++ \cite{guo2025snapmogen} & 16.90 & 8.93 & 10.57 & 22.62 & 19.66 & 9.06 & 11.34 & 20.39 & 16.22 & 8.16 & 9.96 & 37.88 \\
    SOKE (mBART)~\cite{zuo2025soke} & 14.34 & 7.38 & 9.68 & 23.64 & 14.98 & 7.75 & 10.08 & 16.12 & 12.54 & 6.04 & 7.72 & 34.30 \\
    SOKE (Qwen3)~\cite{zuo2025soke} & 14.05 & 7.29 & 9.59 & 25.05 & 14.43 & 7.18 & 10.09 & 18.48 & 11.88 & 6.09 & 7.65 & 38.53 \\
    SignSparK$^\dagger$~\cite{low2026signspark} & 12.13 & 6.08 & 9.95 & 36.84 & 12.84 & 6.24 & 10.76 & 32.48 & 9.28 & 4.88 & 6.73 & 58.76 \\
    MaDiS~\cite{zuo2026madis} & 11.93 & 5.99 & 7.96 & 38.61 & 12.59 & 6.59 & 9.45 & 34.27 & 9.40 & 4.97 & 6.59 & 57.94 \\

    \midrule
    SEEL (Ours) & \textbf{10.86} & \textbf{5.48} & \textbf{7.36} & \textbf{41.84} & \textbf{10.93} & \textbf{5.62} & \textbf{8.88} & \textbf{36.05} & \textbf{8.47} & \textbf{4.45} & \textbf{6.19} & \textbf{62.46} \\
    
    \bottomrule
    \end{tabular}
    \vspace{-2mm}
    \caption{Comparison with state-of-the-art gloss-free sign language generation methods. DTW-JPE (All) scores are obtained using official code where available and our re-implementations of~\cite{baltatzis2024nsa, yin-etal-2024-t2s}. $^*$DTW-JPE (Body/Hands) scores are reported in~\cite{zuo2025soke}. $^\dagger$Reproduced on the SOKE-fitted data~\cite{zuo2025soke} using the official code without keyframe inputs for a fair comparison.}
    \label{tab:sota}
    \vspace{-2mm}
\end{table*}

\begin{figure*}[t]
    \centering
    \includegraphics[width=1.0\linewidth]{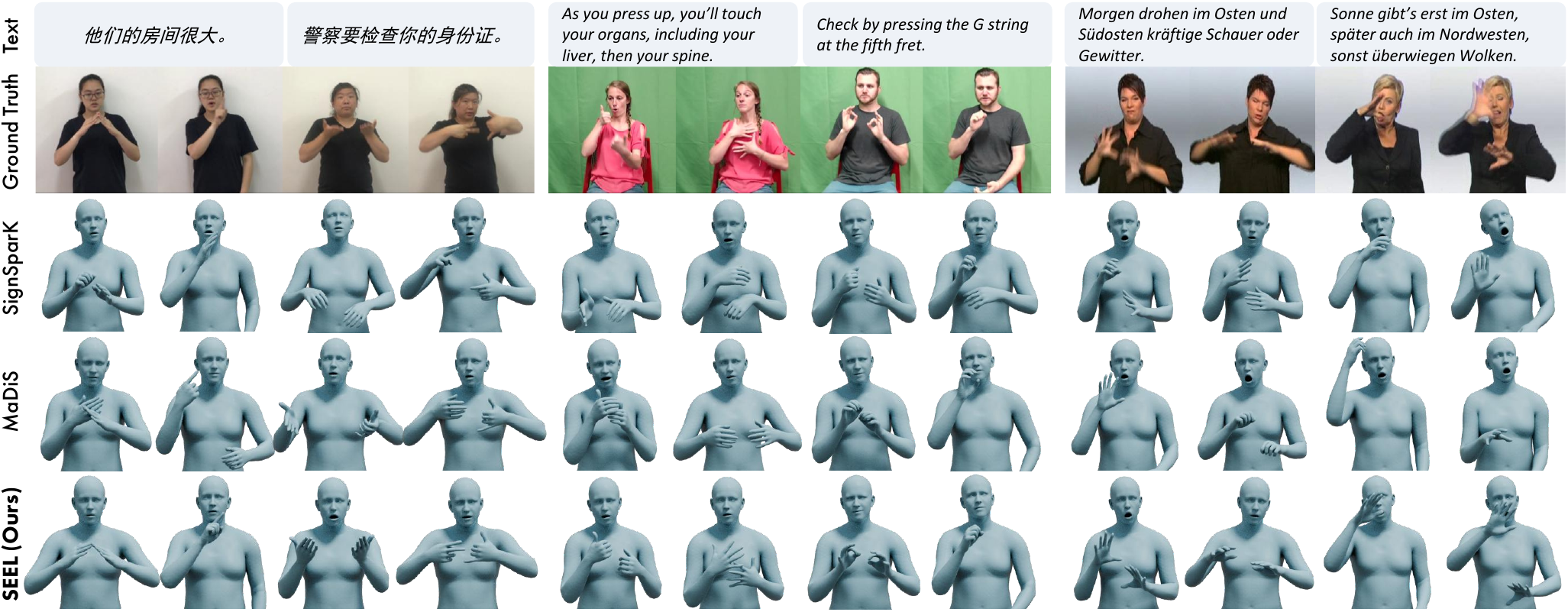}
    \vspace{-7mm}
    \caption{Qualitative comparisons of generated signs between our proposed method, SEEL, and the SOTA methods, SignSparK~\cite{low2026signspark} and MaDiS~\cite{zuo2026madis}, on the test sets of CSL-Daily (left), How2Sign (middle), and Phoenix-2014T (right).}
    \label{fig:qual}
    \vspace{-3mm}
\end{figure*}

\section{Experiments}
\noindent\textbf{Datasets.} 
We evaluate SEEL on three widely-adopted sign language benchmarks: CSL-Daily~\cite{csl-daily}, How2Sign~\cite{duarte2021how2sign}, and Phoenix-2014T~\cite{2014T}, covering Chinese, American, and German Sign Languages, respectively. We use the curated SMPL-X poses provided in~\cite{zuo2025soke} as motion representations.

\noindent\textbf{Evaluation Metrics.} 
To account for temporal length variations between generated and ground-truth sign motions, we follow~\cite{zuo2025soke, low2026signspark, baltatzis2024nsa} and adopt dynamic time warping over joint position error (DTW-JPE) as our primary metric for sequence-level 3D motion accuracy. 
Note that we do not adopt Procrustes alignment, as aligning rotation yields overoptimistic errors~\cite{black2023bedlam}. 
Beyond evaluating the upper body and hands separately, we also report pelvis-relative whole-body error across body, hand, and facial keypoints, denoted as DTW-JPE (All), to capture holistic motion fidelity. 
To assess semantic alignment between generated motions and input text, we report R@1 retrieval accuracy~\cite{bensabath2025text} using SiCLIP~\cite{zuo2026madis} as the embedding extractor. 
For tokenization, we measure motion reconstruction error via whole-body mean per joint position error (MPJPE) and evaluate codebook utilization via codebook perplexity~\cite{vqvae} across both reconstruction ($\text{PPL}_{\mathrm{rec}}$) and generation ($\text{PPL}_{\mathrm{gen}}$). 
Both perplexity metrics are computed as the geometric mean across the body, left-hand, and right-hand codebooks (of sizes 96, 192, and 192), yielding a theoretical maximum of 152.4.

\begin{figure*}[t]
\centering
\begin{minipage}[b]{0.36\textwidth}
    \centering
    \includegraphics[width=1.0\linewidth]{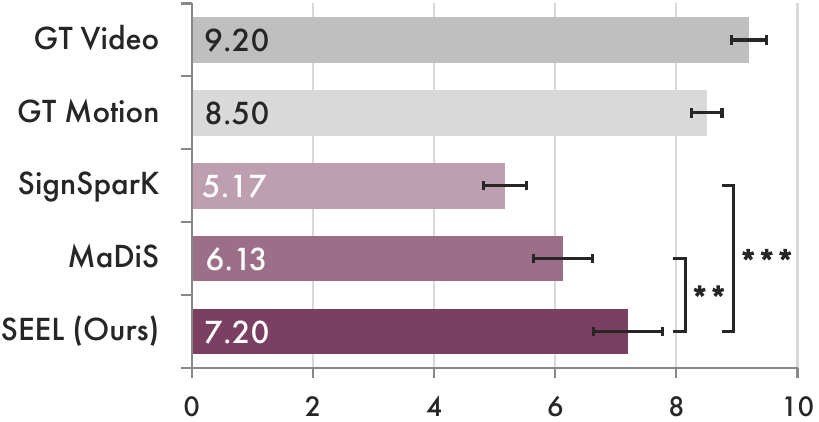}
    \vspace{-6mm}
    \captionof{figure}{User study with professional CSL signers. We report average ratings of ground truth, prior SOTA~\cite{low2026signspark,zuo2026madis}, and SEEL. Error bars indicate 95\% confidence intervals ($^{**}p < 0.01$, $^{***}p < 0.001$, Wilcoxon signed-rank test~\cite{Wilcoxon_test}).}
    \label{fig:user}
\end{minipage}%
\hfill
\begin{minipage}[b]{0.62\textwidth}
    \centering
    \small
    \setlength{\tabcolsep}{2pt}
    \resizebox{\linewidth}{!}{%
    \begin{tabular}{l|cc|ccc|cccc}
    \toprule
    \multirow{3}{*}{Training Paradigm} & \multirow{3}{*}{REG} & \multirow{3}{*}{GET} & \multicolumn{3}{c|}{Tokenization} & \multicolumn{4}{c}{Generation} \\
    \cmidrule(lr){4-6} \cmidrule(lr){7-10}
     & & & \multirow{2}{*}{MPJPE$\downarrow$} & \multirow{2}{*}{$\text{PPL}_{\mathrm{rec}}$$\uparrow$} & \multirow{2}{*}{$\text{PPL}_{\mathrm{gen}}$$\uparrow$} & \multicolumn{3}{c}{DTW-JPE$\downarrow$} & SiCLIP$\uparrow$ \\
    \cmidrule(lr){7-9} \cmidrule(lr){10-10}
     & & & & & & All & Body & Hands & R@1 \\
    \midrule
    Two-Stage Baseline & -- & -- & 43.54 & 87.84 & 68.90 & 13.81 & 6.93 & 9.27 & 31.49 \\
    \midrule
    Na\"ive E2E Baseline & -- & -- & 45.95 & 66.56 & 53.21 & 14.41 & 7.48 & 9.85 & 30.58 \\
    SEEL (REG Only) & \checkmark & -- & 43.54 & 87.84 & 82.98 & 12.38 & 6.22 & 8.17 & 36.64 \\
    SEEL (GET Only) & -- & \checkmark & 40.19 & 133.07 & 106.54 & 11.73 & 6.05 & 7.98 & 37.47 \\
    SEEL (Ours) & \checkmark & \checkmark & \textbf{38.62} & \textbf{140.17} & \textbf{121.16} & \textbf{10.86} & \textbf{5.48} & \textbf{7.36} & \textbf{41.84} \\
    \bottomrule
    \end{tabular}%
    }
    \captionof{table}{Ablation study on training paradigms and core components of SEEL. We evaluate motion reconstruction quality, codebook utilization via reconstruction and generation perplexity, and downstream generation performance. (E2E: End-to-End.)}
    \label{tab:ablation_main}
\end{minipage}
\vspace{-2mm}
\end{figure*}

\noindent\textbf{Implementation Details.} 
We adopt the masked diffusion language model from~\cite{zuo2026madis} as our generator, and use the same motion tokenizer architecture (encoder, decoder, and codebook) as~\cite{zuo2025soke, zuo2026madis}. 
For REG, each candidate set $\mathcal{K}_t$ contains 8 tokens: the reference token $c_t$ plus 7 sampled candidates in a 4:2:1 ratio (generator top-$k$, nearest codebook neighbors, and random codes), with the filtering margin $\delta = 1.0$\,mm.
We empirically set the temperatures $\tau_G = 1.0$ and $\tau_E = 0.8$. 
To stabilize early training, the loss weights ($\lambda_{\mathrm{reg}}$, $\lambda_{\mathrm{get}}$, $\lambda_{\mathrm{motion}}$) are set to zero for the first 5 epochs and then linearly warmed up over the next 5 epochs to 1.0.
Without pretraining on sign language data, we jointly train the randomly initialized motion tokenizer and the Qwen3-0.6B-Base~\cite{yang2025qwen3} initialized generator for 150 epochs on each dataset, using the AdamW~\cite{adamw} optimizer with a batch size of 64 per GPU, an initial learning rate of $2\times 10^{-4}$, and cosine learning rate decay. 
At inference, we generate motion tokens over 25 unmasking steps~\cite{zuo2026madis}. All models are trained on 2x H200 GPUs. Further details, training/inference costs, and hyperparameter analyses are in the appendix.

\begin{figure}[t]
\centering
\includegraphics[width=1.0\linewidth]{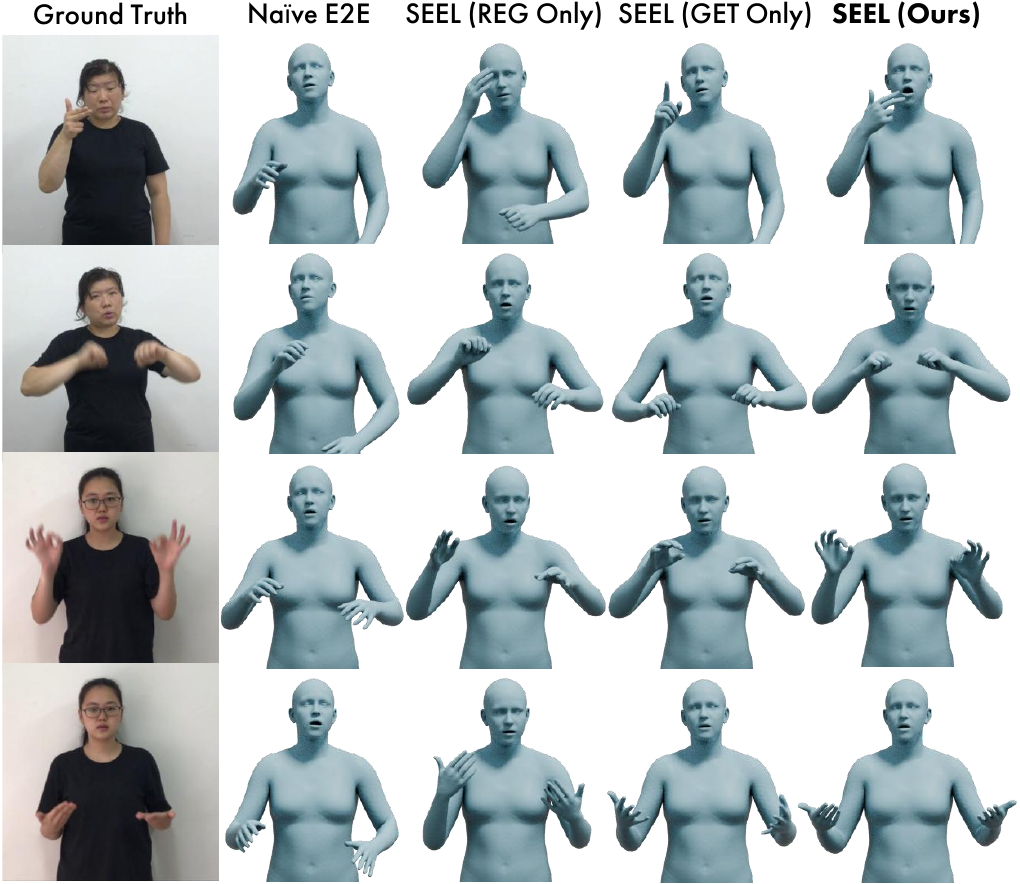}
\vspace{-7mm}
\caption{Qualitative ablation of SEEL components.}
\vspace{-4mm}
\label{fig:abl_qual}
\end{figure}

\subsection{Comparison with State-of-the-Art Methods}
\noindent\textbf{Quantitative Comparison.}
\cref{tab:sota} compares SEEL against leading gloss-free SLG methods. 
Among recent approaches, two-stage language-model frameworks such as SOKE~\cite{zuo2025soke} and MaDiS~\cite{zuo2026madis} remain limited by a frozen, separately trained tokenizer and stage-wise optimization. By jointly optimizing tokenization and generation with REG and GET, SEEL consistently improves upon both methods, surpassing MaDiS without requiring additional pretraining.
Furthermore, SEEL outperforms the alternative generation paradigm SignSparK~\cite{low2026signspark}, which introduces conditional flow matching with reconstruction regularization. Overall, SEEL achieves SOTA performance across all three benchmarks and all evaluation metrics.

\noindent\textbf{Qualitative Comparison.}
\cref{fig:qual} provides a visual comparison of SEEL against leading SOTA methods~\cite{low2026signspark,zuo2026madis}, complementing the benchmark results in \cref{tab:sota}. Across diverse signing examples, SEEL produces 3D motions that match the ground truth more closely than competing approaches, exhibiting more accurate body pose, arm positioning, handshapes, and more precise finger articulations. Additional video demonstrations are available on the project page.

\noindent\textbf{User Study with Professional Signers.}
To assess the intelligibility of the generated signs, we conducted a user study with 6 professional CSL signers (5+ years of signing experience), who rated the sign accuracy and semantic alignment with text annotations on a scale from 1 to 10. Specifically, participants evaluated 25 randomly shuffled generations from SignSparK, MaDiS, and our proposed SEEL, alongside the corresponding ground-truth motions and videos for reference. All participants were compensated above the average local hourly rate. As shown in \cref{fig:user}, SEEL achieves an average rating of 7.20, substantially outperforming SignSparK (5.17) and MaDiS (6.13).

\subsection{Ablation Study}
Following prior works~\cite{zuo2025soke,lee2026signer,zuo2026madis}, we conduct all ablation studies on CSL-Daily unless otherwise specified.

\noindent\textbf{Effect of Core Components.}
\cref{tab:ablation_main} ablates our training paradigms and core modules using the MaDiS~\cite{zuo2026madis} backbone under the standard two-stage recipe without additional pretraining. 
Even with EMA updates and codebook resets~\cite{jiang2023motiongpt,zuo2025soke,t2m-gpt}, the two-stage pipeline underutilizes its frozen codebook, while na\"ive E2E training with STE degrades the codebook and worsens both reconstruction and generation. 
In contrast, REG alone leverages soft targets to better exploit the codebook, whereas GET alone expands active codebook capacity and refines reconstruction.
Combining both in full SEEL maximizes codebook utilization and achieves the best performance. Qualitatively (\cref{fig:abl_qual}), SEEL produces more accurate arm placement, handshapes, and finger articulation than other variants.

\begin{table}[t]
\small
\setlength\tabcolsep{3pt}
\centering
\resizebox{\columnwidth}{!}{%
\begin{tabular}{l|cccc}
\toprule
\multirow{2}{*}{Supervision Mechanism} & \multicolumn{3}{c}{DTW-JPE$\downarrow$} & SiCLIP$\uparrow$ \\
\cmidrule(lr){2-4} \cmidrule(lr){5-5}
 & All & Body & Hands & R@1 \\
\midrule
$\mathcal{L}_{\mathrm{motion}}$ via STE (Na\"ive Baseline) & 14.41 & 7.48 & 9.85 & 30.58 \\
$\mathcal{L}_{\mathrm{motion}}$ via Codebook Expectations & 13.42 & 6.82 & 9.05 & 32.54 \\
Vanilla Label Smoothing~\cite{label_smooth} & 13.73 & 6.89 & 8.97 & 32.05 \\
Motion-Aware One-Hot Target & 13.04 & 6.67 & 8.81 & 33.16 \\
Motion-Aware Soft Target (Ours) & \textbf{12.38} & \textbf{6.22} & \textbf{8.17} & \textbf{36.64} \\
\bottomrule
\end{tabular}%
}
\vspace{-3mm}
\caption{Ablation study on supervision mechanisms in REG.}
\label{tab:ablation_reg_sup}
\vspace{-2mm}
\end{table}

\noindent\textbf{Effect of Supervision Mechanism in REG.}
\cref{tab:ablation_reg_sup} compares generator supervision mechanisms in REG (with $\lambda_{\mathrm{get}}=0$). Directly backpropagating $\mathcal{L}_{\mathrm{motion}}$ via STE suffers from biased gradient estimation and yields poor generation quality. 
Replacing STE with codebook expectations ($\tilde{\mathbf{c}}$) circumvents gradient bias, yet supervising the generator via a single motion sequence remains less informative than explicitly scoring competing token candidates.
Within token-space supervision, vanilla label smoothing ($\epsilon = 0.1$) distributes probability mass uniformly without awareness of motion fidelity, while a motion-aware one-hot target selects the lowest-error candidate but ignores kinematic similarities among other plausible tokens. In contrast, our motion-aware soft target weights candidates by their decoded motion errors, achieving the strongest generation performance.

\begin{table}[t]
\small
\setlength\tabcolsep{4pt}
\centering
\resizebox{\columnwidth}{!}{%
\begin{tabular}{ccc|cccc}
\toprule
\multicolumn{3}{c|}{Candidate Source} & \multicolumn{3}{c}{DTW-JPE$\downarrow$} & SiCLIP$\uparrow$ \\
\cmidrule(lr){1-3} \cmidrule(lr){4-6} \cmidrule(lr){7-7}
Top-$k$ & Neighbors & Random & All & Body & Hands & R@1 \\
\midrule
-- & -- & \checkmark & 13.55 & 7.14 & 9.42 & 32.99 \\
-- & \checkmark & -- & 13.10 & 6.89 & 9.10 & 34.86 \\
\checkmark & -- & -- & 12.91 & 6.74 & 8.97 & 35.32 \\
\checkmark & \checkmark & -- & 12.53 & 6.36 & 8.39 & 36.23 \\
\checkmark & \checkmark & \checkmark & \textbf{12.38} & \textbf{6.22} & \textbf{8.17} & \textbf{36.64} \\
\bottomrule
\end{tabular}
}
\vspace{-3mm}
\caption{Ablation study on candidate source composition in REG.}
\label{tab:ablation_can}
\vspace{-1mm}
\end{table}

\begin{table}[t]
\small
\setlength\tabcolsep{2pt}
\centering
\resizebox{\columnwidth}{!}{%
\begin{tabular}{l|c|cccc}
\toprule
\multirow{2}{*}{Gradient Propagation} & \multirow{2}{*}{MPJPE$\downarrow$} & \multicolumn{3}{c}{DTW-JPE$\downarrow$} & SiCLIP$\uparrow$ \\
\cmidrule(lr){3-5} \cmidrule(lr){6-6}
& & All & Body & Hands & R@1 \\
\midrule
Straight-Through (STE) & 45.95 & 14.41 & 7.48 & 9.85 & 30.58 \\
Gumbel-Softmax~\cite{gumbel_softmax} & 45.28 & 14.33 & 7.36 & 9.60 & 31.47 \\
GET ($E \to G$ Only) & 41.40 & 13.24 & 6.96 & 9.11 & 34.69 \\
GET ($G \to D$ Only) & 42.69 & 12.96 & 6.89 & 8.78 & 35.04 \\
GET (Ours) & \textbf{40.19} & \textbf{11.73} & \textbf{6.05} & \textbf{7.98} & \textbf{37.47} \\
\bottomrule
\end{tabular}%
}
\vspace{-3mm}
\caption{Ablation study on gradient propagation in GET.}
\label{tab:ablation_get}
\vspace{-2mm}
\end{table}

\noindent\textbf{Effect of Candidate Sources in REG.}
\cref{tab:ablation_can} ablates the candidate sources for soft target construction in REG. Among single sources, random sampling performs worst due to sparse coverage of relevant tokens, codebook neighbors provide stronger local alternatives, and the generator's top-$k$ predictions perform best by directly calibrating likely outputs. Combining top-$k$ predictions with neighbors brings clear gains, and incorporating random tokens further boosts performance through codebook exploration.

\begin{figure}[t]
\centering
\includegraphics[width=1.0\linewidth]{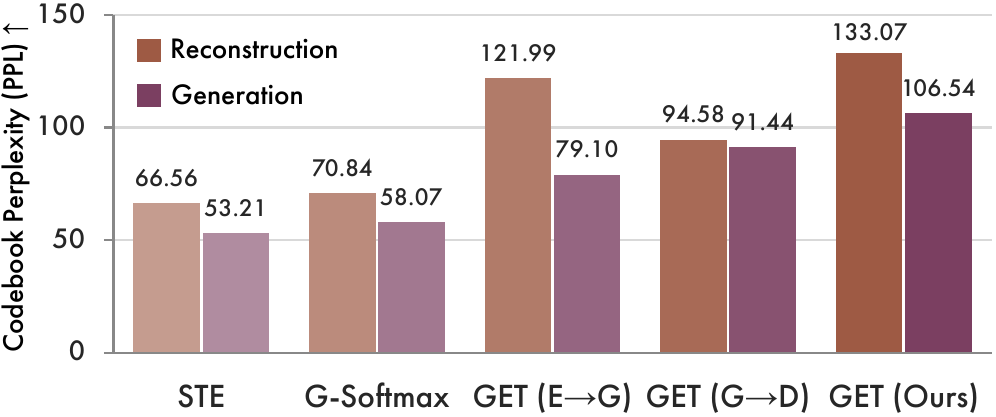}
\vspace{-6mm}
\caption{Comparison of codebook perplexity across different gradient propagation mechanisms. (G-Softmax: Gumbel-Softmax.)} 
\vspace{-1mm}
\label{fig:ppl_get}
\end{figure}

\noindent\textbf{Effect of Gradient Propagation in GET.}
\cref{tab:ablation_get} and \cref{fig:ppl_get} evaluate various gradient propagation mechanisms across the two interfaces (with $\lambda_{\mathrm{reg}}=0$). 
Both STE and Gumbel-Softmax suffer from low perplexity due to sparse hard assignments with biased gradients or high sampling variance~\cite{tucker2017rebar}, degrading reconstruction and generation. 
In contrast, our continuous codebook expectations reveal complementary roles when routing generation feedback to $E$ (using STE at the remaining interface in single-interface variants): bridging $E \rightarrow G$ alone raises $\text{PPL}_{\mathrm{rec}}$ via soft neighborhood assignments but leaves $\text{PPL}_{\mathrm{gen}}$ limited (79.10), confirming that utilization regularization alone is insufficient without $G \rightarrow D$ feedback through $p_t^G$ (boosting $\text{PPL}_{\mathrm{gen}}$ to 91.44, and 106.54 when combined).
Bridging both interfaces simultaneously eliminates STE across the entire pathway, achieving the best performance across all metrics.

\begin{table}[t]
\small
\setlength\tabcolsep{3pt}
\centering
\begin{tabular}{l|c|cccc}
\toprule
\multirow{2}{*}{Method} & \multirow{2}{*}{MPJPE$\downarrow$} & \multicolumn{3}{c}{DTW-JPE$\downarrow$} & SiCLIP$\uparrow$ \\
\cmidrule(lr){3-5} \cmidrule(lr){6-6}
& & All & Body & Hands & R@1 \\
\midrule
SOKE (Qwen3)~\cite{zuo2025soke} & 43.54 & 14.05 & 7.29 & 9.59 & 25.05 \\
+ REG Only & 43.54 & 12.17 & 6.35 & 8.54 & 33.90 \\
+ GET Only & 40.77 & 11.86 & 6.12 & 8.33 & 35.09 \\
+ SEEL (Ours) & \textbf{39.09} & \textbf{11.14} & \textbf{5.75} & \textbf{7.64} & \textbf{39.27} \\
\bottomrule
\end{tabular}%
\vspace{-2mm}
\caption{Generalization of SEEL to autoregressive generation.}
\label{tab:abl_ar}
\vspace{-3mm}
\end{table}

\noindent\textbf{Generalization to Autoregressive Generation.}
To verify that SEEL generalizes beyond masked diffusion language models, \cref{tab:abl_ar} applies REG and GET to an autoregressive generator, SOKE (Qwen3)~\cite{zuo2025soke}. Specifically, under causal teacher forcing, we replace randomly masked contexts with causal prefix contexts, scoring next-token candidates in REG while feeding continuous prefix embeddings to the generator in GET. Consistent with our findings on the masked generator, both mechanisms bring clear individual gains, and combining them substantially improves both reconstruction and generation.
\section{Conclusion}

We present SEEL, an end-to-end framework that jointly optimizes a motion tokenizer and language-model generator for 3D sign language generation. SEEL resolves the objective mismatch of decoupled pipelines while avoiding the gradient bias and codebook collapse of na\"ive joint training. Reconstruction-aware generation decodes candidate tokens in context to guide the generator with motion-aware soft targets, while generation-aware tokenization routes generated-motion gradients through codebook expectations to optimize the encoder. Extensive evaluations show that SEEL achieves SOTA performance across three benchmarks.

\section*{LLM Usage Statement}
Large language models (specifically, Gemini, ChatGPT, and Claude) were used exclusively for language polishing, grammatical error correction, and improving the clarity and readability of the author-drafted text across the manuscript. No LLMs or generative AI tools were used for research ideation, methodological formulation, code implementation, experimental design, data analysis, or result generation. The authors have thoroughly reviewed and verified all revised text and take full responsibility for the integrity, accuracy, and originality of this work.

\section*{Acknowledgements}
S.~Zafeiriou and part of this research were funded by the EPSRC Project GNOMON (EP/X011364/1) and a Turing AI Fellowship (EP/Z534699/1).
F.~Ni was supported by the European Union's Horizon Europe research and innovation programme under the Marie Sk{\l}odowska-Curie grant agreement No.~101282325.
The authors thank Yuecong Min for coordinating the user study and the anonymous signers who participated in it.

The authors acknowledge the use of resources provided by the Isambard-AI National AI Research Resource (AIRR). Isambard-AI~\cite{mcintosh2024isambard} is operated by the University of Bristol and is funded by the UK Government's Department for Business, Innovation, Science and Trade via UK Research and Innovation; and the Science and Technology Facilities Council [ST/AIRR/I-A-I/1023].
The authors acknowledge the use of resources provided by the Zenith supercomputer, operated by the University of Cambridge and delivered in partnership with Dell Technologies and AMD. Zenith is funded by UK Research and Innovation (UKRI) under the UK AI Research Resource (AIRR; grant ST/Z000890/1), an initiative of the UK Government's Department for Business, Innovation, Science, and Trade, and the National Compute Resource (NCR; grant UKRI4064) programmes.
The authors also acknowledge computational resources and support provided by the Imperial College Research Computing Service (DOI:~10.14469/hpc/2232).

{
    \small
    \bibliographystyle{ieeenat_fullname}
    \bibliography{main}

@String(IJCV = {Int. J. Comput. Vis.})

@String(CVPR= {IEEE Conf. Comput. Vis. Pattern Recog.})

@String(ICCV= {Int. Conf. Comput. Vis.})

@String(ECCV= {Eur. Conf. Comput. Vis.})

@String(BMVC= {Brit. Mach. Vis. Conf.})

@String(ICLR = {Int. Conf. Learn. Represent.})

@String(AAAI = {AAAI})

@String(IJCV  = {IJCV})

@String(CVPR  = {CVPR})

@String(ICCV  = {ICCV})

@String(ECCV  = {ECCV})

@String(BMVC  =	{BMVC})

@String(ICLR  = {ICLR})

@inproceedings{2014T,
  title={Neural Sign Language Translation},
  author={Necati Cihan Camgoz  and  Simon Hadfield  and  Oscar Koller  and  Hermann Ney and  Richard Bowden},
  booktitle={CVPR},
  year={2018},
}

@inproceedings{csl-daily,
  title={Improving Sign Language Translation with Monolingual Data by Sign Back-Translation},
  author={Zhou, Hao and Zhou, Wengang and Qi, Weizhen and Pu, Junfu and Li, Houqiang},
  booktitle={CVPR},
  year={2021}
}

@inproceedings{label_smooth,
  title={Rethinking the inception architecture for computer vision},
  author={Szegedy, Christian and Vanhoucke, Vincent and Ioffe, Sergey and Shlens, Jon and Wojna, Zbigniew},
  booktitle={CVPR},
  year={2016}
}

@article{dtw,
  title={Dynamic time warping},
  author={M{\"u}ller, Meinard},
  journal={Information retrieval for music and motion},
  pages={69--84},
  year={2007},
  publisher={Springer}
}

@article{mbart,
  title={Multilingual denoising pre-training for neural machine translation},
  author={Liu, Yinhan and Gu, Jiatao and Goyal, Naman and Li, Xian and Edunov, Sergey and Ghazvininejad, Marjan and Lewis, Mike and Zettlemoyer, Luke},
  journal={TACL},
  volume={8},
  pages={726--742},
  year={2020},
}

@inproceedings{zuo2024simple,
  title={A Simple Baseline for Spoken Language to Sign Language Translation with 3D Avatars},
  author={Ronglai Zuo and Fangyun Wei and Zenggui Chen and Brian Mak and Jiaolong Yang and Xin Tong},
  booktitle={ECCV},
  year={2024}
}

@inproceedings{arkushin2023ham2pose,
  title={Ham2Pose: Animating Sign Language Notation Into Pose Sequences},
  author={Arkushin, Rotem Shalev and Moryossef, Amit and Fried, Ohad},
  booktitle={CVPR},
  pages={21046--21056},
  year={2023}
}

@inproceedings{yin2021including,
  title={Including Signed Languages in Natural Language Processing},
  author={Yin, Kayo and Moryossef, Amit and Hochgesang, Julie and Goldberg, Yoav and Alikhani, Malihe},
  booktitle={ACL},
  pages={7347--7360},
  year={2021}
}

@article{stoll2020text2sign,
  title={Text2Sign: towards sign language production using neural machine translation and generative adversarial networks},
  author={Stoll, Stephanie and Camgoz, Necati Cihan and Hadfield, Simon and Bowden, Richard},
  journal={IJCV},
  volume={128},
  number={4},
  pages={891--908},
  year={2020},
}

@inproceedings{huang2022dualsign,
  title={DualSign: Semi-Supervised Sign Language Production with Balanced Multi-Modal Multi-Task Dual Transformation},
  author={Huang, Wencan and Zhao, Zhou and He, Jinzheng and Zhang, Mingmin},
  booktitle={ACM MM},
  year={2022}
}

@inproceedings{saunders2022signing,
  title={Signing at scale: Learning to co-articulate signs for large-scale photo-realistic sign language production},
  author={Saunders, Ben and Camgoz, Necati Cihan and Bowden, Richard},
  booktitle={CVPR},
  pages={5141--5151},
  year={2022}
}

@inproceedings{saunders2020progressive,
  title={Progressive transformers for end-to-end sign language production},
  author={Saunders, Ben and Camgoz, Necati Cihan and Bowden, Richard},
  booktitle={ECCV},
  pages={687--705},
  year={2020},
}

@article{saunders2021continuous,
  title={Continuous {3D} multi-channel sign language production via progressive transformers and mixture density networks},
  author={Saunders, Ben and Camgoz, Necati Cihan and Bowden, Richard},
  journal={IJCV},
  year={2021},
}

@inproceedings{saunders2021mixed,
  title={Mixed signals: Sign language production via a mixture of motion primitives},
  author={Saunders, Ben and Camgoz, Necati Cihan and Bowden, Richard},
  booktitle={ICCV},
  pages={1919--1929},
  year={2021}
}

@inproceedings{forte2023reconstructing,
  title={Reconstructing Signing Avatars From Video Using Linguistic Priors},
  author={Forte, Maria-Paola and Kulits, Peter and Huang, Chun-Hao P and Choutas, Vasileios and Tzionas, Dimitrios and Kuchenbecker, Katherine J and Black, Michael J},
  booktitle={CVPR},
  pages={12791--12801},
  year={2023}
}

@inproceedings{tang2022gloss,
  title={Gloss semantic-enhanced network with online back-translation for sign language production},
  author={Tang, Shengeng and Hong, Richang and Guo, Dan and Wang, Meng},
  booktitle={ACM MM},
  year={2022}
}

@inproceedings{muller2022considerations,
  title={Considerations for meaningful sign language machine translation based on glosses},
  author={M{\"u}ller, Mathias and Jiang, Zifan and Moryossef, Amit and Gonzales, Annette Rios and Ebling, Sarah},
  booktitle={ACL},
  pages={682--693},
  year={2023}
}

@inproceedings{stoll2022there,
  title={There and back again: 3d sign language generation from text using back-translation},
  author={Stoll, Stephanie and Mustafa, Armin and Guillemaut, Jean-Yves},
  booktitle={3DV},
  pages={187--196},
  year={2022},
}

@inproceedings{baltatzis2024nsa,
  title={Neural Sign Actors: A diffusion model for 3D sign language production from text},
  author={Baltatzis, Vasileios and Potamias, Rolandos Alexandros and Ververas, Evangelos and Sun, Guanxiong and Deng, Jiankang and Zafeiriou, Stefanos},
  booktitle={CVPR},
  pages={1985--1995},
  year={2024}
}

@article{jiang2023motiongpt,
  title={Motiongpt: Human motion as a foreign language},
  author={Jiang, Biao and Chen, Xin and Liu, Wen and Yu, Jingyi and Yu, Gang and Chen, Tao},
  journal={NeurIPS},
  volume={36},
  pages={20067--20079},
  year={2023}
}

@inproceedings{yu2025signavatars,
  title={Signavatars: A large-scale 3d sign language holistic motion dataset and benchmark},
  author={Yu, Zhengdi and Huang, Shaoli and Cheng, Yongkang and Birdal, Tolga},
  booktitle={ECCV},
  pages={1--19},
  year={2024},
}

@inproceedings{duarte2021how2sign,
  title={How2sign: a large-scale multimodal dataset for continuous american sign language},
  author={Duarte, Amanda and Palaskar, Shruti and Ventura, Lucas and Ghadiyaram, Deepti and DeHaan, Kenneth and Metze, Florian and Torres, Jordi and Giro-i-Nieto, Xavier},
  booktitle={CVPR},
  pages={2735--2744},
  year={2021}
}

@inproceedings{
adamw,
title={Decoupled Weight Decay Regularization},
author={Ilya Loshchilov and Frank Hutter},
booktitle={ICLR},
year={2019},
}

@inproceedings{shi2024pose,
  title={Pose-Guided Fine-Grained Sign Language Video Generation},
  author={Shi, Tongkai and Hu, Lianyu and Shang, Fanhua and Feng, Jichao and Liu, Peidong and Feng, Wei},
  booktitle={ECCV},
  pages={392--409},
  year={2024},
}

@inproceedings{yin-etal-2024-t2s,
    title = "{T}2{S}-{GPT}: Dynamic Vector Quantization for Autoregressive Sign Language Production from Text",
    author = "Yin, Aoxiong  and
      Li, Haoyuan  and
      Shen, Kai  and
      Tang, Siliang  and
      Zhuang, Yueting",
    booktitle = "ACL",
    year = "2024",
}

@inproceedings{fg2024signavatar,
  title={SignAvatar: Sign Language 3D Motion Reconstruction and Generation},
  author={Dong, Lu and Chaudhary, Lipisha and Xu, Fei and Wang, Xiao and Lary, Mason and Nwogu, Ifeoma},
  booktitle={FG},
  year={2024}
}

@inproceedings{signgen,
  title={SignGen: End-to-End Sign Language Video Generation with Latent Diffusion},
  author={Fan Qi and Yu Duan and Changsheng Xu and Huaiwen Zhang},
  booktitle={ECCV},
  year={2024}
}

@inproceedings{t2m-gpt,
  title={Generating human motion from textual descriptions with discrete representations},
  author={Zhang, Jianrong and Zhang, Yangsong and Cun, Xiaodong and Zhang, Yong and Zhao, Hongwei and Lu, Hongtao and Shen, Xi and Shan, Ying},
  booktitle={CVPR},
  pages={14730--14740},
  year={2023}
}

@inproceedings{chen2023executing,
  title={Executing your commands via motion diffusion in latent space},
  author={Chen, Xin and Jiang, Biao and Liu, Wen and Huang, Zilong and Fu, Bin and Chen, Tao and Yu, Gang},
  booktitle={CVPR},
  year={2023}
}

@inproceedings{latentdiff,
  title={High-resolution image synthesis with latent diffusion models},
  author={Rombach, Robin and Blattmann, Andreas and Lorenz, Dominik and Esser, Patrick and Ommer, Bj{\"o}rn},
  booktitle={CVPR},
  year={2022}
}

@inproceedings{vqvae,
  title={Neural discrete representation learning},
  author={Van Den Oord, Aaron and Vinyals, Oriol and others},
  booktitle={NeurIPS},
  year={2017}
}

@inproceedings{fang2025signllm,
  title={Signllm: Sign languages production large language models},
  author={Fang, Sen and Wang, Lei and Zheng, Ce and Tian, Yapeng and Chen, Chen},
  booktitle={ICCVW},
  year={2025}
}

@inproceedings{tang2025discrete,
  title={Discrete to Continuous: Generating Smooth Transition Poses from Sign Language Observations},
  author={Tang, Shengeng and He, Jiayi and Cheng, Lechao and Wu, Jingjing and Guo, Dan and Hong, Richang},
  booktitle={CVPR},
  year={2025}
}

@inproceedings{tang2025sign,
  title={Sign-idd: Iconicity disentangled diffusion for sign language production},
  author={Tang, Shengeng and He, Jiayi and Guo, Dan and Wei, Yanyan and Li, Feng and Hong, Richang},
  booktitle={AAAI},
  year={2025}
}

@inproceedings{zuo2025soke,
    title={Signs as Tokens: A Retrieval-Enhanced Multilingual Sign Language Generator},
    author={Zuo, Ronglai and Potamias, Rolandos Alexandros and Ververas, Evangelos and Deng, Jiankang and Zafeiriou, Stefanos},
    booktitle={ICCV},
    year={2025}
}

@inproceedings{guo2025snapmogen,
  title={SnapMoGen: Human Motion Generation from Expressive Texts},
  author={Guo, Chuan and Hwang, Inwoo and Wang, Jian and Zhou, Bing},
  booktitle={NeurIPS},
  year={2025}
}

@inproceedings{guo2024momask,
  title={Momask: Generative masked modeling of 3d human motions},
  author={Guo, Chuan and Mu, Yuxuan and Javed, Muhammad Gohar and Wang, Sen and Cheng, Li},
  booktitle={CVPR},
  year={2024}
}

@InProceedings{Shi_2025_ICCV,
    author    = {Shi, Tongkai and Hu, Lianyu and Shang, Fanhua and Gao, Liqing and Feng, Wei},
    title     = {GReg: Geometry-Aware Region Refinement for Sign Language Video Generation},
    booktitle = {ICCV},
    year      = {2025},
}

@article{llada,
  title={Large language diffusion models},
  author={Nie, Shen and Zhu, Fengqi and You, Zebin and Zhang, Xiaolu and Ou, Jingyang and Hu, Jun and Zhou, Jun and Lin, Yankai and Wen, Ji-Rong and Li, Chongxuan},
  journal={NeurIPS},
  year={2025}
}

@article{yang2025qwen3,
  title={Qwen3 technical report},
  author={Yang, An and Li, Anfeng and Yang, Baosong and Zhang, Beichen and Hui, Binyuan and Zheng, Bo and Yu, Bowen and Gao, Chang and Huang, Chengen and Lv, Chenxu and others},
  journal={arXiv preprint arXiv:2505.09388},
  year={2025}
}

@inproceedings{bensabath2025text,
  title={Text-driven 3d hand motion generation from sign language data},
  author={Bensabath, L{\'e}ore and Petrovich, Mathis and Varol, G{\"u}l},
  booktitle={CVPR},
  year={2026}
}

@inproceedings{black2023bedlam,
  title={Bedlam: A synthetic dataset of bodies exhibiting detailed lifelike animated motion},
  author={Black, Michael J and Patel, Priyanka and Tesch, Joachim and Yang, Jinlong},
  booktitle={CVPR},
  year={2023}
}

@inproceedings{chang2022maskgit,
  title={Maskgit: Masked generative image transformer},
  author={Chang, Huiwen and Zhang, Han and Jiang, Lu and Liu, Ce and Freeman, William T},
  booktitle={CVPR},
  year={2022}
}

@inproceedings{fang2026stable,
  title={Stable Signer: Hierarchical Sign Language Generative Model},
  author={Fang, Sen and Feng, Yalin and Zhong, Hongbin and Zhang, Yanxin and Metaxas, Dimitris N},
  booktitle={ACL},
  year={2026}
}

@article{xia2026signmask,
  title={SignMask: Structure-aware Masked Modeling for Holistic 3D Sign Language Production},
  author={Xia, Yibo and Zhan, Qihui and Luo, Xiaoyan and Shi, Xiaofeng and Wang, Yunhong},
  journal={ACM TOMM},
  year={2026},
}

@incollection{mcintosh2024isambard,
  title={Isambard-ai: a leadership-class supercomputer optimised specifically for artificial intelligence},
  author={McIntosh-Smith, Simon and Alam, Sadaf and Woods, Christopher},
  booktitle={Proceedings of the Cray User Group},
  year={2024},
  publisher={ACM}
}

@article{zuo2026madis,
  title={MaDiS: Taming Masked Diffusion Language Models for Sign Language Generation},
  author={Zuo, Ronglai and Potamias, Rolandos Alexandros and Sun, Qi and Ververas, Evangelos and Deng, Jiankang and Zafeiriou, Stefanos},
  journal={arXiv preprint arXiv:2601.19577},
  year={2026}
}

@article{bengio2013ste,
  title={Estimating or propagating gradients through stochastic neurons for conditional computation},
  author={Bengio, Yoshua and L{\'e}onard, Nicholas and Courville, Aaron},
  journal={arXiv preprint arXiv:1308.3432},
  year={2013}
}

@inproceedings{low2026signspark,
  title={SignSparK: efficient multilingual sign language production via sparse keyframe learning},
  author={Low, Jianhe and Symeonidis-Herzig, Alexandre and Ivashechkin, Maksym and Sincan, Ozge Mercanoglu and Bowden, Richard},
  booktitle={ECCV},
  year={2026},
}

@inproceedings{lee2026signer,
  title={Signer: Temporally grounded sign language generation via time-resolved conditioning},
  author={Lee, Taeryung and Nam, Hyeongjin and Moon, Gyeongsik and Lee, Kyoung Mu},
  booktitle={ECCV},
  year={2026},
}

@article{wang2026advanced,
  title={Advanced sign language video generation with compressed and quantized multi-condition tokenization},
  author={Wang, Cong and Deng, Zexuan and Jiang, Zhiwei and Yin, Yafeng and Shen, Fei and Cheng, Zifeng and Ge, Shiping and Gan, Shiwei and Gu, Qing},
  journal={NeurIPS},
  year={2026}
}

@inproceedings{liu2026bridging,
  title={Bridging the Gap Between Semantics and Reconstruction: Unifying Sign Language Translation and Production},
  author={Liu, Xiao and Gan, Shiwei and Yin, Yafeng and Yin, Jiaxin and Guo, Bowen and Sun, Yaqi and Jiang, Zhiwei and Xie, Lei},
  booktitle={ACM MM},
  year={2026}
}

@inproceedings{pelykh2026signrefine,
  title={SignRefine: Adapting Foundational Video Models for Sign Language Generation},
  author={Pelykh, Anton and Fish, Edward and Sincan, Ozge Mercanoglu and Bowden, Richard},
  booktitle={ECCV},
  year={2026},
}

@article{qi2026signmod,
  title={SignMoD: Sign Language Video Generation via Mixture of Diffusion},
  author={Qi, Fan and Duan, Yu and Zhang, Huaiwen and Xu, Changsheng},
  journal={IEEE TPAMI},
  year={2026},
}

@inproceedings{liu2026signpr,
  title={SignPR: A Progressive Vector-Quantized Diffusion Framework for Sign Language Production},
  author={Liu, Xiao and Gan, Shiwei and Yin, Yafeng and Guo, Bowen and Jiang, Zhiwei and Meng, Shunmei and Xie, Lei and Lu, Sanglu},
  booktitle={CVPR},
  year={2026}
}

@inproceedings{yu2026focal,
  title={Focal-General Diffusion Model with Semantic Consistent Guidance for Sign Language Production},
  author={Yu, Yiheng and Liu, Sheng and Feng, Yuan and Jin, Zhelun and Jiang, Yining and Xu, Min},
  booktitle={CVPR},
  year={2026}
}

@inproceedings{yu2024magvit,
  title={Language model beats diffusion-tokenizer is key to visual generation},
  author={Yu, Lijun and Lezama, Jos{\'e} and Gundavarapu, Nitesh Bharadwaj and Versari, Luca and Sohn, Kihyuk and Minnen, David and Cheng, Yong and Gupta, Agrim and Gu, Xiuye and Hauptmann, Alexander G and others},
  booktitle={ICLR},
  year={2024}
}

@article{ramanujan2026worse,
  title={When Worse is Better: Navigating the Compression Generation Trade-off In Visual Tokenization},
  author={Ramanujan, Vivek and Tirumala, Kushal and Aghajanyan, Armen and Zettlemoyer, Luke and Farhadi, Ali},
  journal={NeurIPS},
  year={2025}
}

@inproceedings{vqgan,
  title={Taming transformers for high-resolution image synthesis},
  author={Esser, Patrick and Rombach, Robin and Ommer, Bjorn},
  booktitle={CVPR},
  year={2021}
}

@inproceedings{fsq,
  title={Finite scalar quantization: Vq-vae made simple},
  author={Mentzer, Fabian and Minnen, David and Agustsson, Eirikur and Tschannen, Michael},
  booktitle={ICLR},
  year={2024}
}

@inproceedings{yin2019understanding,
  title={Understanding straight-through estimator in training activation quantized neural nets},
  author={Yin, Penghang and Lyu, Jiancheng and Zhang, Shuai and Osher, Stanley and Qi, Yingyong and Xin, Jack},
  booktitle={ICLR},
  year={2019}
}

@inproceedings{qiao2026signbot,
  title={Signbot: Learning human-to-humanoid sign language interaction},
  author={Qiao, Guanren and Lin, Sixu and Zuo, Ronglai and Wu, Zhizheng and Jia, Kui and Liu, Guiliang},
  booktitle={ICRA},
  year={2026},
}

@inproceedings{khan2026sign,
  title={From Sign Language Generation to Humanoid Execution: Vision-Language Guided Retargeting with Collision Mitigation},
  author={Khan, Nabeela and Wu, Bowen and Shi, Runwu and Yen, Benjamin and Ashizawa, Takeshi and Ishi, Carlos Toshinori and Minato, Takashi and Nakadai, Kazuhiro},
  booktitle={IROS},
  year={2026}
}

@article{liu2026traceable,
  title={Traceable Human-to-Humanoid Sign Language Benchmarking},
  author={Liu, Ao and Tang, Shengeng and Cheng, Lechao and Hao, Yanbin and Bao, Bingkun and Hong, Richang},
  journal={arXiv preprint arXiv:2609.33354},
  year={2026}
}

@inproceedings{wang2026phosa,
  title={PHOSA: Photorealistic 3D Sign Avatar Modeling and Benchmark},
  author={Wang, Haodong and Hu, Hezhen and Zhou, Wengang and Li, Houqiang},
  booktitle={ECCV},
  year={2026},
}

@inproceedings{zhang2025guava,
  title={Guava: Generalizable upper body 3d gaussian avatar},
  author={Zhang, Dongbin and Liu, Yunfei and Lin, Lijian and Zhu, Ye and Li, Yang and Qin, Minghan and Li, Yu and Wang, Haoqian},
  booktitle={ICCV},
  year={2025},
}

@inproceedings{ye2026hybrid,
  title={Hybrid Autoregressive-Diffusion Model for Real-Time Sign Language Production},
  author={Ye, Maoxiao and Ye, Xinfeng and Manoharan, Sathiamoorthy},
  booktitle={ACL},
  year={2026}
}

@article{ivashechkin2025signsplat,
  title={Signsplat: Rendering sign language via gaussian splatting},
  author={Ivashechkin, Maksym and Mendez, Oscar and Bowden, Richard},
  journal={arXiv preprint arXiv:2505.02108},
  year={2025}
}

@inproceedings{chu2026eostok,
  title={End-to-end autoregressive image generation with 1d semantic tokenizer},
  author={Chu, Wenda and Zhang, Bingliang and Han, Jiaqi and Li, Yizhuo and Yang, Linjie and Yue, Yisong and Guo, Qiushan},
  booktitle={ICML},
  year={2026}
}

@article{zheng2026genfirst,
  title={GenFirst: Generation Before Reconstruction for Stable End-to-End Latent Generative Modeling},
  author={Zheng, Guangting and Zhang, Yiyuan and Yang, Tao and Chen, Yunpeng and Zhu, Rui and Deng, Jiajun and Zhang, Yanyong},
  journal={arXiv preprint arXiv:2608.29335},
  year={2026}
}

@inproceedings{leng2025repae,
  title={Repa-e: Unlocking vae for end-to-end tuning with latent diffusion transformers},
  author={Leng, Xingjian and Singh, Jaskirat and Hou, Yunzhong and Xing, Zhenchang and Xie, Saining and Zheng, Liang},
  booktitle={ICCV},
  year={2025},
}

@inproceedings{yu2025repa,
  title={Representation alignment for generation: Training diffusion transformers is easier than you think},
  author={Yu, Sihyun and Kwak, Sangkyung and Jang, Huiwon and Jeong, Jongheon and Huang, Jonathan and Shin, Jinwoo and Xie, Saining},
  booktitle={ICLR},
  year={2025}
}

@inproceedings{yao2025reconstruction,
  title={Reconstruction vs. generation: Taming optimization dilemma in latent diffusion models},
  author={Yao, Jingfeng and Yang, Bin and Wang, Xinggang},
  booktitle={CVPR},
  year={2025},
}

@inproceedings{peebles2023scalable,
  title={Scalable diffusion models with transformers},
  author={Peebles, William and Xie, Saining},
  booktitle={ICCV},
  year={2023},
}

@article{sun2024autoregressive,
  title={Autoregressive Model Beats Diffusion: {Llama} for Scalable Image Generation},
  author={Sun, Peize and Jiang, Yi and Chen, Shoufa and Zhang, Shilong and Peng, Bingyue and Luo, Ping and Yuan, Zehuan},
  journal={arXiv preprint arXiv:2406.06525},
  year={2024}
}

@inproceedings{wang2025larp,
  title={Larp: Tokenizing videos with a learned autoregressive generative prior},
  author={Wang, Hanyu and Suri, Saksham and Ren, Yixuan and Chen, Hao and Shrivastava, Abhinav},
  booktitle={ICLR},
  year={2025}
}

@article{zhang2026videorepa,
  title={Videorepa: Learning physics for video generation through relational alignment with foundation models},
  author={Zhang, Xiangdong and Liao, Jiaqi and Zhang, Shaofeng and Meng, Fanqing and Wan, Xiangpeng and Yan, Junchi and Cheng, Yu},
  journal={NeurIPS},
  year={2026}
}

@inproceedings{blattmann2023align,
  title={Align your latents: High-resolution video synthesis with latent diffusion models},
  author={Blattmann, Andreas and Rombach, Robin and Ling, Huan and Dockhorn, Tim and Kim, Seung Wook and Fidler, Sanja and Kreis, Karsten},
  booktitle={CVPR},
  year={2023},
}

@inproceedings{yan2026lgtok,
  title={Language-Guided Transformer Tokenizer for Human Motion Generation},
  author={Yan, Sheng and Wang, Yong and Du, Xin and Yuan, Junsong and Liu, Mengyuan},
  booktitle={ECCV},
  year={2026},
}

@inproceedings{an2026signflip,
  title={SignFLIP: A Unified Model for Sign Language Translation and Generation via Stage-wise Alignment at Scale},
  author={An, Zhaoyi and Tan, Sihan and Hwang, Youngbae and Nakadai, Kazuhiro and Kawakami, Rei},
  booktitle={Findings of EMNLP},
  year={2026}
}

@inproceedings{guo2024unsupervised,
  title={Unsupervised sign language translation and generation},
  author={Guo, Zhengsheng and He, Zhiwei and Jiao, Wenxiang and Wang, Xing and Wang, Rui and Chen, Kehai and Tu, Zhaopeng and Xu, Yong and Zhang, Min},
  booktitle={Findings of ACL},
  year={2024}
}

@article{lakhal2025diverse,
  title={Diverse Signer Avatars with Manual and Non-Manual Feature Modelling for Sign Language Production},
  author={Lakhal, Mohamed Ilyes and Bowden, Richard},
  journal={arXiv preprint arXiv:2508.15988},
  year={2025}
}

@book{Wilcoxon_test,
  title={Practical nonparametric statistics},
  volume={493},
  author={W. J. Conover },
  year={1999},
  publisher={John Wiley \& Sons}
}

@inproceedings{huh2023straightening,
  title={Straightening out the straight-through estimator: Overcoming optimization challenges in vector quantized networks},
  author={Huh, Minyoung and Cheung, Brian and Agrawal, Pulkit and Isola, Phillip},
  booktitle={ICML},
  year={2023},
}

@inproceedings{gumbel_softmax,
  title={Categorical reparameterization with gumbel-softmax},
  author={Jang, Eric and Gu, Shixiang and Poole, Ben},
  booktitle={ICLR},
  year={2017}
}

@article{tucker2017rebar,
  title={Rebar: Low-variance, unbiased gradient estimates for discrete latent variable models},
  author={Tucker, George and Mnih, Andriy and Maddison, Chris J and Lawson, John and Sohl-Dickstein, Jascha},
  journal={NeurIPS},
  year={2017}
}

@inproceedings{symeonidis2026m3t,
  title={M3T: Discrete Multi-Modal Motion Tokens for Sign Language Production},
  author={Symeonidis-Herzig, Alexandre and Low, Jianhe and Sincan, Ozge Mercanoglu and Bowden, Richard},
  booktitle={BMVC},
  year={2026}
}
}

\setcounter{table}{0}\renewcommand{\thetable}{S\arabic{table}}
\setcounter{figure}{0}\renewcommand{\thefigure}{S\arabic{figure}}
\setcounter{equation}{0}\renewcommand{\theequation}{S\arabic{equation}}

\appendix
\section{More Implementation Details}

\subsection{Extended Method and Architecture Details}
\noindent\textbf{Decoupled Motion Tokenizer.}
Following~\cite{zuo2025soke,zuo2026madis}, we adopt an open-source three-stream 1D convolutional VQ-VAE~\cite{zuo2025soke} that independently encodes the upper body (including the face), left hand, and right hand. Each encoder comprises 2 downsampling stages, each containing a stride-2 convolution followed by 3 residual blocks with hidden width 512. The resulting temporal downsampling factor is 4x, producing latent features $\mathbf{z}^{r}=E^r(\mathbf{x}^{r})\in\mathbb{R}^{T\times512}$ for each part $r$. The corresponding decoders use residual blocks and nearest-neighbor upsampling to restore the original temporal resolution. The codebooks contain 96, 192, and 192 entries for the body, left hand, and right hand, respectively. Codebook entries are updated using exponential moving averages (EMA) with decay $\gamma=0.99$. At each training-batch update, entries whose EMA assignment count falls below 1.0 are reinitialized from current encoder features~\cite{jiang2023motiongpt,t2m-gpt,zuo2025soke}. The tokenizer objective is
\begin{equation}
    \mathcal{L}_{\mathrm{tok}}
    =
    \|\hat{\mathbf{x}}-\mathbf{x}\|_1
    +
    \beta_{\mathrm{cmt}}
    \sum_{r}
    \mathrm{MSE}\!\left(
        \mathbf{z}^{r},
        \mathrm{sg}(\mathbf{C}^{r}[\mathbf{c}^{r}])
    \right),
\end{equation}
where $\beta_{\mathrm{cmt}}=0.02$ weights the commitment loss,
which encourages encoder features to remain close to their
assigned code embeddings.

\noindent\textbf{Masked Diffusion Language Model (MDLM).}
Our generator $G$ follows~\cite{zuo2026madis} and adapts Qwen3-0.6B-Base~\cite{yang2025qwen3} into a bidirectional masked diffusion language model~\cite{llada} by replacing causal self-attention with full bidirectional attention over motion positions. Given written text $y$ tokenized by the pretrained Qwen3 text tokenizer, the three parallel 512-d motion token streams are linearly projected into the 1024-d LM hidden space and combined per temporal step via mixture-of-parts fusion~\cite{zuo2026madis}, then decoded through three stream-specific prediction heads over their respective vocabularies. During training, a masking ratio $\alpha \sim \mathcal{U}(0, 1)$ is sampled per sequence, and each motion position is independently masked to form the prediction set $\mathcal{S}$ while keeping unmasked context tokens $\mathbf{c}_{\mathrm{ctx}}$ visible. Target sequence length is modeled via an end-of-sentence token. 
At inference, generation starts from a fully masked sequence of 100 motion tokens (corresponding to at most 400 frames) and performs 25 iterative steps using UTC~\cite{zuo2026madis}, unmasking 4 tokens per step by retaining the highest-confidence predictions while remasking the remaining positions for subsequent refinement.
We set the sampling temperature to zero to ensure reproducibility.

\noindent\textbf{Gumbel-Softmax Baseline.} 
For the Gumbel-Softmax baseline in Tab.~5 and Fig.~8 of the main paper, we apply straight-through Gumbel-Softmax~\cite{gumbel_softmax} at both quantization interfaces ($E \to G$ over negative squared code distances and $G \to D$ over generator logits). 
Following~\cite{gumbel_softmax}, we exponentially anneal the temperature from 1.0 to 0.5 via $\tau = \max(0.5, \exp(-\mu s))$ at training step $s$ with $\mu = 3 \times 10^{-5}$, while keeping all other training settings identical to GET.

\subsection{Datasets and Evaluation Protocol}

\noindent\textbf{Datasets.}
We evaluate SEEL on three widely adopted sign language benchmarks: CSL-Daily~\cite{csl-daily}, How2Sign~\cite{duarte2021how2sign}, and Phoenix-2014T~\cite{2014T}, containing 20K, 35K, and 8K video-text pairs for Chinese, American, and German sign languages, respectively. We follow the official train/validation/test splits and adopt the curated SMPL-X poses~\cite{zuo2025soke} as our motion representation, where each frame consists of 133 parameters: 3D rotations for 11 upper-body joints and 30 hand joints, together with 10 facial expression parameters.

\noindent\textbf{Evaluation Metrics.}
We evaluate both tokenization and generation quality using the following metrics:
\begin{itemize}
    \item \textbf{Reconstruction Quality.} MPJPE measures the mean per-joint position error between reconstructed and ground-truth 3D motions. 
    
    \item \textbf{Codebook Utilization.} For each body-part stream $r \in \{\text{Body}, \text{L-Hand}, \text{R-Hand}\}$, we compute the part-wise perplexity $\text{PPL}^{r} = \exp(H^{r})$, where $H^{r} = -\sum_k p_k^{r} \log p_k^{r}$ is the entropy of its empirical token frequencies. Intuitively, $\text{PPL}^{r}$ reflects the effective number of active codes: a higher PPL indicates broader, more uniform codebook utilization, whereas a lower PPL indicates that token usage is concentrated on a smaller subset of codes (\eg, codebook collapse). We aggregate across the three streams via the geometric mean:
    \begin{equation}
        \text{PPL} = \left(\prod_{r} \text{PPL}^{r}\right)^{\!\frac{1}{3}} = \exp\!\left(\frac{1}{3}\sum_{r} H^{r}\right),
    \end{equation}
    evaluated during reconstruction ($\text{PPL}_{\text{rec}}$) and generation ($\text{PPL}_{\text{gen}}$). We use the geometric rather than the arithmetic mean because perplexity operates on an exponential scale: the geometric mean reflects the average part-wise entropy, prevents streams with larger vocabularies (hands vs.\ body) from dominating the score, and remains sensitive to codebook collapse in any stream.

    \item \textbf{Generation Quality.} We evaluate geometric fidelity using DTW-JPE, which computes the average 3D joint position error after dynamic time warping~\cite{dtw} alignment between generated and ground-truth motion sequences, reported overall (All) as well as separately for the upper body (Body) and hands (Hands). We also evaluate semantic alignment using SiCLIP (R@1), which measures Top-1 sign-to-text retrieval accuracy with the open-source pretrained sign-text contrastive evaluator~\cite{zuo2026madis}, reflecting how accurately generated signs convey the input text.
\end{itemize}

\section{More Quantitative Results}
\begin{table}[t]
\centering
\small
\setlength{\tabcolsep}{2pt}
\resizebox{\columnwidth}{!}{%
\begin{tabular}{l|cc|cc}
\toprule
\multirow{2}{*}{Training Paradigm} & \multicolumn{2}{c|}{Training Cost} & \multicolumn{2}{c}{Generation Performance} \\
\cmidrule(lr){2-3} \cmidrule(lr){4-5}
& GPUh$\downarrow$ & GPU Mem. (GB)$\downarrow$ & DTW-JPE (All)$\downarrow$ & SiCLIP (R@1)$\uparrow$ \\
\midrule
Two-Stage Baseline$^*$ & 38.15 & 56.59 & 13.81 & 31.49 \\
Na\"ive E2E Baseline & 40.96 & 63.27 & 14.41 & 30.58 \\
SEEL (REG Only) & 44.05 & 60.58 & 12.38 & 36.64 \\
SEEL (GET Only) & 58.92 & 93.19 & 11.73 & 37.47 \\
\midrule
SEEL (Ours) & 66.20 & 96.17 & 10.86 & 41.84 \\
\bottomrule
\end{tabular}
}
\caption{Training cost and generation performance across paradigms on CSL-Daily. All models are trained to convergence on 2x H200 GPUs with a batch size of 64 per GPU. $^*$Two-Stage GPUh sums Stage-1 tokenizer and Stage-2 generator training, with peak memory across both stages reported. Because encoder $E$, REG, and GET are discarded after training, all paradigms share identical inference efficiency, achieving a throughput of 135.33 tokens/s (510.54 FPS) on a single GPU over 25 unmasking steps.}
\label{tab:efficiency}
\end{table}

\subsection{Training Cost and Inference Efficiency}
\cref{tab:efficiency} reports the training cost and generation quality across all training paradigms on CSL-Daily.
Adding REG alone introduces only a modest overhead over the two-stage baseline (+15.5\% GPU hours, with lower peak memory than Na\"ive E2E because batched candidate decoding runs without gradients), while already noticeably improving DTW-JPE.
GET accounts for most of the additional compute and memory due to its extra forward-backward pass through the frozen generator $G$ and decoder $D$ with active input gradients.
Overall, full SEEL incurs a 1.74x training-time overhead to reduce DTW-JPE by 21.4\% and improve SiCLIP R@1 by 10.35 points.
Crucially, because the encoder $E$, REG, and GET are used exclusively during training, SEEL introduces \textbf{zero inference overhead}: all five paradigms share identical model parameters and achieve the same inference speed (510.54 FPS on a single GPU).

\begin{table}[t]
\centering
\small
\setlength{\tabcolsep}{2pt}
\resizebox{\columnwidth}{!}{%
\begin{tabular}{l|cc|ccc|cccc}
    \toprule
    \multirow{3}{*}{Training Paradigm} & \multirow{3}{*}{REG} & \multirow{3}{*}{GET} & \multicolumn{3}{c|}{Tokenization} & \multicolumn{4}{c}{Generation} \\
    \cmidrule(lr){4-6} \cmidrule(lr){7-10}
     & & & \multirow{2}{*}{MPJPE$\downarrow$} & \multirow{2}{*}{$\text{PPL}_{\mathrm{rec}}$$\uparrow$} & \multirow{2}{*}{$\text{PPL}_{\mathrm{gen}}$$\uparrow$} & \multicolumn{3}{c}{DTW-JPE$\downarrow$} & SiCLIP$\uparrow$ \\
    \cmidrule(lr){7-9} \cmidrule(lr){10-10}
     & & & & & & All & Body & Hands & R@1 \\
    \midrule
    Two-Stage Baseline & -- & -- & 44.56 & 94.42 & 80.63 & 13.56 & 6.93 & 11.04 & 27.21 \\
    \midrule
    Na\"ive E2E Baseline & -- & -- & 46.39 & 78.86 & 63.24 & 14.43 & 7.46 & 11.77 & 26.67 \\
    SEEL (REG Only) & \checkmark & -- & 44.56 & 94.42 & 90.89 & 12.50 & 6.52 & 10.15 & 31.81 \\
    SEEL (GET Only) & -- & \checkmark & 41.81 & 135.84 & 120.56 & 11.86 & 6.12 & 9.41 & 34.20 \\
    SEEL (Ours) & \checkmark & \checkmark & \textbf{41.03} & \textbf{140.42} & \textbf{126.35} & \textbf{10.93} & \textbf{5.62} & \textbf{8.88} & \textbf{36.05} \\
    \bottomrule
    \end{tabular}%
}
\caption{Ablation study on training paradigms and core components of SEEL on How2Sign.}
\label{tab:abl_h2s}
\end{table}

\subsection{More Ablation Studies}

\noindent\textbf{Cross-Dataset Component Ablations.}
To verify that the synergy between REG and GET generalizes across sign languages and dataset scales, ~\cref{tab:abl_h2s,tab:abl_pho} replicate our paradigm ablation on How2Sign and Phoenix-2014T. Both benchmarks consistently reproduce the trends observed on CSL-Daily. Na\"ive E2E suffers from codebook collapse and underperforms the two-stage baseline in both tokenization and generation. 
Enabling REG alone leaves tokenizer reconstruction unchanged while steadily narrowing the gap between reconstruction and generation perplexity and lifting both geometric fidelity and semantic alignment. By shaping encoder representations for generator predictability, GET likewise improves generation quality while also enhancing reconstruction fidelity and codebook utilization. Combining both mechanisms in full SEEL achieves the strongest tokenization and generation performance across every metric on both benchmarks.

\begin{table}[t]
\centering
\small
\setlength{\tabcolsep}{2pt}
\resizebox{\columnwidth}{!}{%
\begin{tabular}{l|cc|ccc|cccc}
    \toprule
    \multirow{3}{*}{Training Paradigm} & \multirow{3}{*}{REG} & \multirow{3}{*}{GET} & \multicolumn{3}{c|}{Tokenization} & \multicolumn{4}{c}{Generation} \\
    \cmidrule(lr){4-6} \cmidrule(lr){7-10}
     & & & \multirow{2}{*}{MPJPE$\downarrow$} & \multirow{2}{*}{$\text{PPL}_{\mathrm{rec}}$$\uparrow$} & \multirow{2}{*}{$\text{PPL}_{\mathrm{gen}}$$\uparrow$} & \multicolumn{3}{c}{DTW-JPE$\downarrow$} & SiCLIP$\uparrow$ \\
    \cmidrule(lr){7-9} \cmidrule(lr){10-10}
     & & & & & & All & Body & Hands & R@1 \\
    \midrule
    Two-Stage Baseline & -- & -- & 45.77 & 92.95 & 77.50 & 10.32 & 5.44 & 7.64 & 51.81 \\
    \midrule
    Na\"ive E2E Baseline & -- & -- & 47.96 & 78.13 & 60.94 & 10.90 & 5.75 & 8.03 & 49.42 \\
    SEEL (REG Only) & \checkmark & -- & 45.77 & 92.95 & 88.63 & 9.63 & 5.25 & 7.13 & 56.68 \\
    SEEL (GET Only) & -- & \checkmark & 42.91 & 136.71 & 113.64 & 9.18 & 4.97 & 6.60 & 58.45 \\
    SEEL (Ours) & \checkmark & \checkmark & \textbf{42.04} & \textbf{141.72} & \textbf{120.02} & \textbf{8.47} & \textbf{4.45} & \textbf{6.19} & \textbf{62.46} \\
    \bottomrule
    \end{tabular}%
}
\caption{Ablation study on training paradigms and core components of SEEL on Phoenix-2014T.}
\label{tab:abl_pho}
\end{table}

\begin{table}[t]
\centering
\small
\setlength{\tabcolsep}{3pt}
\resizebox{\columnwidth}{!}{%
\begin{tabular}{l|ccc|ccc}
    \toprule
    \multirow{2}{*}{Training Paradigm} & \multicolumn{3}{c|}{$\text{PPL}_{\mathrm{rec}}$$\uparrow$} & \multicolumn{3}{c}{$\text{PPL}_{\mathrm{gen}}$$\uparrow$} \\
    \cmidrule(lr){2-4} \cmidrule(lr){5-7}
    & Body & L-Hand & R-Hand & Body & L-Hand & R-Hand \\
    \midrule
    Two-Stage Baseline & 75.05 & 70.01 & 129.00 & 59.39 & 49.05 & 112.27 \\
    \midrule
    Na\"ive E2E Baseline & 62.39 & 48.35 & 97.75 & 46.63 & 38.47 & 83.99 \\
    SEEL (REG Only) & 75.05 & 70.01 & 129.00 & 66.27 & 67.54 & 127.66 \\
    SEEL (GET Only) & 87.91 & 154.43 & 173.57 & 64.50 & 131.84 & 142.21 \\
    SEEL (Ours) & \textbf{90.91} & \textbf{167.55} & \textbf{180.82} & \textbf{69.57} & \textbf{156.58} & \textbf{163.27} \\
    \bottomrule
    \end{tabular}%
}
\caption{Part-wise codebook perplexity on CSL-Daily. Codebook sizes are 96 for the body and 192 for each hand.}
\label{tab:part_ppl}
\end{table}

\noindent\textbf{Part-Wise Codebook Utilization.}
~\cref{tab:part_ppl} reports part-wise codebook perplexity on CSL-Daily. Under the two-stage baseline, the body stream is reasonably well utilized, whereas both hand codebooks, especially the non-dominant left hand, suffer severe under-utilization. Because single-handed signs are predominantly right-handed, resting poses dominate the left-hand distribution; hard quantization and one-hot cross-entropy amplify this skew into codebook collapse and an oversized left-right gap. Na\"ive E2E worsens this collapse across all streams. In contrast, REG alone preserves baseline reconstruction perplexity while sharply narrowing the reconstruction-generation gap, whereas GET drives the primary hand-codebook recovery by preventing frequent resting poses from taking up most of the codebook.
Full SEEL achieves the highest utilization across every stream, substantially reducing the left-right gap while retaining a modest right-hand advantage reflecting natural linguistic dominance.

\begin{table}[t]
\centering
\small
\begin{tabular}{cc|cccc}
\toprule
\multirow{2}{*}{Space} & \multirow{2}{*}{Distribution} & \multicolumn{3}{c}{DTW-JPE$\downarrow$} & SiCLIP$\uparrow$\\
\cmidrule(lr){3-5}\cmidrule(lr){6-6}
&  & All & Body & Hands & R@1 \\
\midrule
Latent & Softmax & 13.07 & 6.59 & 8.86 & 33.90 \\
Motion & Uniform & 12.89 & 6.44 & 8.51 & 35.14\\
\midrule
Motion & Softmax & \textbf{12.38} & \textbf{6.22} & \textbf{8.17} & \textbf{36.64} \\
\bottomrule
\end{tabular}
\caption{Comparison of various target formulations in REG on CSL-Daily.}
\label{tab:abl_reg_target}
\end{table}

\noindent\textbf{Target Formulations in REG.}
Extending Tab.~3 of the main paper, \cref{tab:abl_reg_target} ablates the target space and distribution in REG over the same candidate set $\mathcal{K}_t$ (without GET) on CSL-Daily. Compared with our motion-space target (Motion, Softmax), computing softmax weights from latent-space distances (Latent, Softmax) clearly degrades performance across all metrics, as Euclidean distance in codebook space ignores decoder non-linearities and temporal context. Similarly, replacing softmax weights with a uniform distribution over the candidates (Motion, Uniform) yields performance inferior to that of our (Motion, Softmax) formulation, showing that equal weighting loses fine-grained ranking among valid candidates.

\begin{table}[t]
\centering
\small
\setlength{\tabcolsep}{2pt}
\resizebox{\columnwidth}{!}{%
\begin{tabular}{l|ccc|cccc}
    \toprule
    \multirow{3}{*}{Method} & \multicolumn{3}{c|}{Tokenization} & \multicolumn{4}{c}{Generation} \\
    \cmidrule(lr){2-4} \cmidrule(lr){5-8}
     & \multirow{2}{*}{MPJPE$\downarrow$} & \multirow{2}{*}{$\text{PPL}_{\mathrm{rec}}$$\uparrow$} & \multirow{2}{*}{$\text{PPL}_{\mathrm{gen}}$$\uparrow$} & \multicolumn{3}{c}{DTW-JPE$\downarrow$} & SiCLIP$\uparrow$ \\
    \cmidrule(lr){5-7} \cmidrule(lr){8-8}
     & & & & All & Body & Hands & R@1 \\
    \midrule
    FSQ + MDLM & 42.36 & 121.80 & 98.74 & 13.37 & 6.82 & 8.83 & 33.15 \\
    \midrule
    w/ REG Only & 42.36 & 121.80 & 113.52 & 12.17 & 6.14 & 8.09 & 37.96 \\
    w/ GET Only & 40.69 & 127.14 & 111.48 & 11.60 & 6.03 & 7.79 & 38.74 \\
    w/ SEEL (Ours) & \textbf{39.42} & \textbf{130.59} & \textbf{125.85} & \textbf{10.92} & \textbf{5.56} & \textbf{7.43} & \textbf{41.92} \\
    \bottomrule
    \end{tabular}%
}
\caption{Generalization of SEEL to finite scalar quantization (FSQ)~\cite{fsq} on CSL-Daily.}
\label{tab:abl_fsq}
\end{table}

\noindent\textbf{Generalization to Finite Scalar Quantization (FSQ).}
Recent sign language generation work such as M3T~\cite{symeonidis2026m3t} adopts FSQ~\cite{fsq} to avoid codebook collapse. To verify that SEEL is not tied to learnable VQ-VAE codebooks, we replace vector quantization with FSQ on CSL-Daily. Specifically, we project each stream's encoder features to three channels and apply channel-wise bounding and rounding with levels [6, 4, 4] for the body and [8, 6, 4] for each hand. This matches our VQ-VAE vocabulary sizes (96 and 192, respectively) so the MDLM architecture remains unchanged. We materialize the resulting Cartesian grid vertices as a fixed implicit codebook $\mathcal{G} \in \mathbb{R}^{N \times 3}$, where each vertex maps one-to-one to a discrete token index and is linearly projected back to the decoder input dimension (512-d). Adapting SEEL requires minimal changes: REG operates identically by looking up the fixed grid vertices for candidate indices $\mathcal{K}_t$ and decoding them to score motion errors; GET computes the soft-assignment distribution $q_t^E$ from the squared Euclidean distances between the bounded pre-rounding latent and all vertices in $\mathcal{G}$, and computes its codebook expectations directly over the fixed grid vertices. 
As in our VQ-VAE formulation, GET gradients update only the encoder and its pre-quantization projection layer.

As shown in \cref{tab:abl_fsq}, two-stage FSQ achieves much higher reconstruction perplexity than two-stage VQ-VAE by avoiding dead codes, yet it still suffers a clear drop in generation perplexity and trails SEEL widely in generation quality. This confirms that preventing codebook collapse alone does not resolve the mismatched objectives between reconstruction and generation. Adding REG preserves baseline reconstruction while sharply narrowing the perplexity gap and improving motion fidelity. Adding GET outperforms REG by shaping encoder features toward generator-predictable grid cells. Combining both components in full SEEL achieves the best results across all metrics, confirming that SEEL generalizes well across various quantization methods.

\subsection{Hyperparameter Sensitivity Analysis}
\begin{table}[t]
\centering
\small
\setlength{\tabcolsep}{3pt}
\begin{tabular}{l|cccccc}
\toprule
$\tau_{G}$ & 0.2 & 0.4 & 0.6 & 0.8 & 1.0 & 1.2 \\
\midrule
DTW-JPE (All)$\downarrow$ & 12.94 & 12.66 & 12.52 & 12.43 & \textbf{12.38} & 12.47 \\
\bottomrule
\end{tabular}
\caption{Sensitivity analysis on $\tau_{G}$ on CSL-Daily.}
\label{tab:hyp_tau_G}
\end{table}

\begin{table}[t]
\centering
\small
\setlength{\tabcolsep}{3pt}
\begin{tabular}{l|cccccc}
\toprule
$\tau_{E}$ & 0.2 & 0.4 & 0.6 & 0.8 & 1.0 & 1.2 \\
\midrule
DTW-JPE (All)$\downarrow$ & 12.30 & 11.99 & 11.85 & \textbf{11.73} & 11.86 & 12.07 \\
\bottomrule
\end{tabular}
\caption{Sensitivity analysis on $\tau_{E}$ on CSL-Daily.}
\label{tab:hyp_tau_E}
\end{table}

\begin{table}[t]
\centering
\small
\begin{tabular}{l|cccc}
\toprule
$|\mathcal{K}_t|$ & 1+3 & 1+7 & 1+14 & 1+28 \\
\midrule
DTW-JPE (All)$\downarrow$ & 12.69 & 12.38 & 12.41 & 12.36 \\
Training Overhead$\downarrow$ & 1.09x & 1.15x & 1.28x & 1.54x \\
\bottomrule
\end{tabular}
\caption{Sensitivity analysis on the candidate set size ($|\mathcal{K}_t|$) in REG on CSL-Daily. In ``1+$X$'', 1 denotes the reference token, which is always included in $\mathcal{K}_t$, and $X$ denotes the remaining candidates drawn from three sources (top-$k$, nearest-neighbor, and random). We allocate these sources in a 1:1:1 ratio for ``1+3'' so that each source contributes one candidate, and use a 4:2:1 ratio by default for all other settings. Training overhead is relative to the two-stage baseline.}
\label{tab:hyp_cand_size}
\end{table}

\noindent\textbf{REG Target Temperature ($\tau_G$).}
\cref{tab:hyp_tau_G} evaluates the sensitivity of REG (without GET) to $\tau_G$ on CSL-Daily. Small values sharpen the target distribution toward hard one-hot supervision, limiting the benefit of soft targets, whereas overly large values flatten the distribution toward uniform weighting and weaken pose ranking. The best performance is achieved when $\tau_G = 1.0$.

\noindent\textbf{GET Assignment Temperature ($\tau_E$).}
\cref{tab:hyp_tau_E} analyzes $\tau_E$ under GET (without REG) on CSL-Daily. Very small temperatures push soft assignments toward hard quantization, restricting gradient flow across neighboring codes, while large temperatures spread weights over distant codes and dilute encoder updates. Setting $\tau_E = 0.8$ achieves the best performance.

\noindent\textbf{Candidate Set Size ($|\mathcal{K}_t|$).}
\cref{tab:hyp_cand_size} analyzes the sensitivity of REG to the candidate set size $|\mathcal{K}_t|$, expressed as ``1+$X$'', where 1 is the always-included reference token and $X$ is the number of additional candidates sampled from three sources: top-$k$ predictions, codebook nearest neighbors, and random tokens. Guided by our single-source ablation in the main paper (Tab.~4), where top-$k$ performs best, nearest neighbors rank second, and random tokens perform worst, we allocate the additional candidates in a 4:2:1 ratio to prioritize each source proportionally to its effectiveness, while ``1+3'' uses a 1:1:1 split to retain one token from each source. 
Expanding $|\mathcal{K}_t|$ from ``1+3'' to ``1+7'' improves DTW-JPE from 12.69 to 12.38 with only a minor increase in training overhead (1.09x to 1.15x). Scaling further to ``1+14'' or ``1+28'' yields saturated performance at noticeably higher training cost. We therefore adopt ``1+7'' as our default setting for the best balance between generation quality and training efficiency.

\section{More Qualitative Results}
The demo videos on our project page present side-by-side comparisons of SEEL against SignSparK~\cite{low2026signspark}, MaDiS~\cite{zuo2026madis}, and the ground-truth RGB videos from the CSL-Daily, How2Sign, and Phoenix-2014T test sets. Across all three benchmarks, SEEL generates more accurate body poses and smoother upper-body motion, while preserving clearer handshapes and finger articulation. Moreover, our generated sequences match ground-truth durations more closely.

\section{Discussion}
\noindent\textbf{Scope of ``End-to-End'' Sign Language Generation.}
The term ``end-to-end'' has been used with different scopes across sign language generation (SLG) works. 
Early continuous regression approaches, such as Progressive Transformers~\cite{saunders2020progressive} and its multi-channel MDN extension~\cite{saunders2021continuous}, use ``end-to-end'' to denote gloss-free translation from spoken text to continuous 3D pose sequences; although trained in a single stage, direct continuous regression still suffers from autoregressive drift and under-articulation even with mitigations such as noise augmentation, adversarial training, or mixture density modeling.
More recently, text-to-video approaches such as SignGen~\cite{signgen} use ``end-to-end'' at the pipeline level to bypass an intermediate pose-generation stage, while still relying on a separately pretrained latent visual autoencoder. 

In contrast, modern discrete-token SLG frameworks avoid continuous regression artifacts by introducing a motion tokenizer, yet remain strictly two-stage at the optimization level because the tokenizer is frozen prior to generator training. 
Accordingly, our claim of end-to-end SLG specifically refers to \emph{end-to-end discrete-token SLG}: jointly optimizing the motion tokenizer and generative language model via our proposed REG and GET to resolve the reconstruction-generation mismatch while avoiding direct continuous frame regression.

\noindent\textbf{Limitations and Future Work.}
Our current work has two main limitations. First, we only validate the effectiveness of SEEL on three sign languages (CSL, ASL, and DGS), leaving broader coverage of additional sign languages and domains unexplored. Second, SEEL generates 3D SMPL-X motion sequences rather than photorealistic avatars directly. In the future, we plan to extend SEEL to more sign languages and combine it with downstream avatar renderers to produce lifelike signing videos.

\noindent\textbf{Broader Impact.}
Sign language generation plays a vital role in enabling bidirectional communication between the deaf and hearing communities. By improving the motion fidelity and semantic alignment of text-to-sign synthesis, SEEL can serve as a backbone for accessible communication systems, educational platforms, and assistive digital services, helping bridge communication barriers and promote inclusiveness.


\end{document}